\documentclass[journal]{IEEEtran}
\usepackage{hyperref}
\usepackage{url}
\usepackage{epsfig}
\usepackage{epstopdf}
\usepackage{cite}
\usepackage{caption}

\ifCLASSINFOpdf
\else
\fi

\begin{document}
\title{Reconciling saliency and object center-bias hypotheses in explaining free-viewing fixations}

\author{Ali Borji and James Tanner

\thanks{ A. Borji is with the Computer Science Department, University of Wisconsin - Milwaukee, 
3200 N. Cramer St., Milwaukee, WI 53211. E-mail: borji@uwm.edu}
\thanks{ J. Tanner is with the Department of Computer Science, University of Southern California, 3641 Watt Way, Los Angeles, CA 90089. E-mail: jetanner@usc.edu}
\thanks{Manuscript received October, 2014.}}

%
%

\markboth{IEEE Transactions on Neural Networks and Learning Systems,~Vol. XXX, No. XXX, XXXXX~2014}%
{Shell \MakeLowercase{\textit{et al.}}: Bare Demo of IEEEtran.cls for Journals}
\maketitle

\begin{abstract}
Predicting where people look in natural scenes has attracted a lot of interest in computer vision and computational neuroscience over the past two decades. Two seemingly contrasting categories of cues have been proposed to influence where people look: \textit{low-level image saliency} and \textit{high-level semantic information}. Our first contribution is to take a detailed look at these cues to confirm the hypothesis proposed by Henderson~\cite{henderson1993eye} and Nuthmann \& Henderson~\cite{nuthmann2010object} that observers tend to look at the center of objects. We analyzed fixation data for scene free-viewing over 17 observers on 60 fully annotated images with various types of objects. Images contained different types of scenes, such as natural scenes, line drawings, and 3D rendered scenes. Our second contribution is to propose a simple combined model of low-level saliency and object center-bias that outperforms each individual component significantly over our data, as well as on the OSIE dataset by Xu et al.~\cite{xu2014predicting}. The results reconcile saliency with object center-bias hypotheses and highlight that both types of cues are important in guiding fixations. Our work opens new directions to understand strategies that humans use in observing scenes and objects, and demonstrates the construction of combined models of low-level saliency and high-level object-based information. 

\end{abstract}


\IEEEpeerreviewmaketitle


\section{Introduction}

\IEEEPARstart{E}YE movements are proxies for overt visual attention. They help us understand how humans and animals allocate their perceptual and cognitive resources towards a limited portion of the observed visual data. They also inform us about characteristics of the filtered data. Understanding and modeling human attentional behavior has become increasingly important recently for two reasons: 1) the abundance of visual data in daily life demands highly efficient filtering methods with low computational complexity, specifically when dealing with natural scenes and videos, and 2) there are many applications in computer vision and robotics such as image/video compression, scene understanding, image thumb-nailing, photo collages, human-robot interaction, and robot localization and navigation that could utilize resource allocation methods.  See~\cite{borji2013state,hayhoe2005eye,itti2001computational,land2001ways,navalpakkam2005modeling,schutz2011eye,tatler2011eye,baluch2011mechanisms,borji2012boosting,borji2014salient} for comprehensive reviews on visual attention.

Where do people look during free viewing of images of natural scenes? A tremendous amount of research in cognitive and computer vision communities has investigated this question for more than a decade, yet it still remains a hot topic~\cite{borji2013state,BorjiTIP}. Two types of cues\footnote{It is not easy to demarcate the category of some cues (e.g., object center-bias, text, face). Some authors have classified cues that influence eye movements into three categories: pixel, object, and semantic. Please see~\cite{xu2014predicting}.
} are believed to influence eye movements in this task: 1) low-level image features (a.k.a., bottom-up visual saliency) such as contrast, edge content, intensity bispectra, color, motion, symmetry, and surprise, and 2) high-level features (i.e., object and semantic information) such as faces and people~\cite{2009Cerf,judd2009learning,humphrey2010potency}, text~\cite{wang2012attraction}, object center priors~\cite{henderson1993eye,nuthmann2010object}, image center priors~\cite{tatler2007central,tseng2009quantifying}, horizontal bias in scene viewing (only a left-ward bias for right handers, no effect for left handers)~\cite{ossandon2014spatial}, semantic object
distances\cite{hwang2011semantic}, scene global context~\cite{2006Torralba}, emotions~\cite{subramanianemotion}, memory~\cite{2005Droll,carmi2006role}, gaze direction~\cite{castelhano2007see,borji2014gaze}, culture~\cite{chua2005cultural}, and survival-related features such as food, sex, danger, pleasure, and
pain~\cite{friston1994value,shen2012top}. Note that, while here we focus on a free-viewing task, some of these factors also play a role in top-down task-driven visual attention~\cite{Yarbus1967,Land1994,Ballard1995,2001Land,Borji_etal13smc,borji2014yarbus,borji2015eyes,Hajimirza2012}.


\subsection{Object center-bias}

As an alternative theory for the hypothesis of image-based saliency (low-level image features, such as
contrast, color, and orientation~\cite{TriesmanGelade,Koch_Ullman85,1998Itti,ReinagelZador99,ParkhustEtal2002,borji2012exploiting}), the object-based hypothesis of attention considers objects as the unit of attention. The latter relates to the \textit{cognitive relevance theory} and the role of cognitive top-down knowledge in attention. According to this theory, objects are manipulated to perform a task (e.g., in sandwich making~\cite{hayhoe2003eye}\footnote{Volunteers were asked to make peanut butter and jelly sandwiches. The participants wore headgear that simultaneously tracked the movement of their eyes and recorded the scene before them.}). Overall, the idea of object-based attention is sensible, as to understand a scene one needs to localize objects,
identify them, and establish their spatial relations. Eye movements tell us how a scene is understood by where they land. There has been some debate whether objects or saliency better predict fixations and the landscape still remains unclear~\cite{Einhauser2008objects,borji2013objects}. Note that object center-bias is different than image center-bias~\cite{tatler2007central}, which is the tendency of observers to preferentially look towards the center of images.

\begin{figure*}[t]
\centering	
\includegraphics[width=14cm,height=11cm]{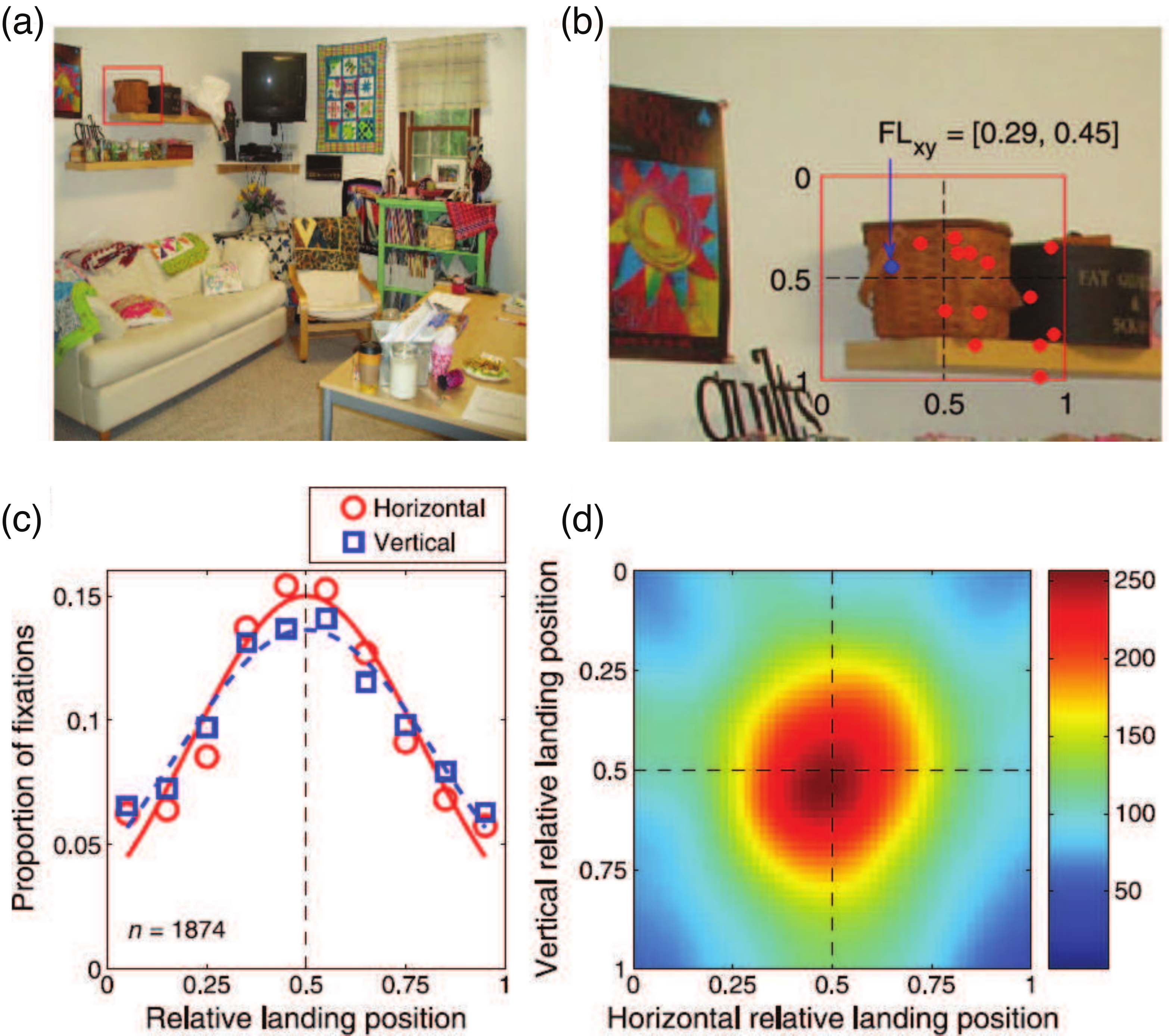} 
\caption{Object-based center-bias. a) An image with a sample annotated object (a basket). Note how loose the bounding box is in this case. b) A close up of the object bounding box and fixations (shown in red). Note that some fixations fall outside the object and on the background. The center of the object is the origin of the coordinate system for fixations. c) Distribution of the horizontal component of landing positions for objects (red circles) and the corresponding distribution of the vertical component of within-object landing positions (blue squares). Circles are data and curves are fitted using truncated Gaussians. The vertical broken line 
indicates the center of the object. Horizontal and vertical lines are overlaid. (d) Corresponding smoothed two-dimensional viewing location histogram. The intersection of the two broken lines marks the center of the object. Images are taken with permission from Nuthmann \& Henderson~\cite{nuthmann2010object}.}
\label{fig:Nuthman}
\end{figure*}

The first fixation-based evidence for object center-bias was demonstrated by Henderson~\cite{henderson1993eye}. He recorded eye movements of observers on line drawings of objects and 
found that viewers' first fixations tended to be near the center of an object, and that
there was a greater tendency to undershoot the center than to overshoot. Later, Trukenbrod and Engbert~\cite{trukenbrod2007oculomotor} reported a similar finding on a serial visual search task.
A more detailed investigation of the object center-bias for objects embedded in naturalistic scenes was conducted by Nuthmann \& Henderson~\cite{nuthmann2010object}\footnote{And also in another recent study~\cite{pajak2013object}.}. These authors measured the fixation landing positions within objects during free viewing of natural scenes, and showed that the \textit{preferred viewing location} (PVL) for real objects in scenes was close to the center of the object (as shown in Figure~\ref{fig:Nuthman}). They also found that when compared to the PVL for real objects, there was less evidence for a PVL for human fixations within saliency proto-objects~\cite{rensink2000dynamic}, identified by an extension to the Itti saliency map model. 
They argued in favor of object-based visual attention and proposed that during naturalistic scene viewing, the eye-movement control system directs eyes in terms of object units. Overall, these findings match with previous findings that observers look at the center of words while reading~\cite{rayner2009rayner}. 
Another piece of evidence comes from a work of Elazary \& Itti~\cite{Elazary_Itti08jov} who showed that objects are usually more salient than the background.

Belardinelli \& Butz~\cite{belardinelligaze} measured the distribution of fixation locations on objects over three tasks: 1) object classification (one of two objects), 2) mimicking lifting an object (lifting task), and
3) mimicking opening an object (opening task). They found that fixations were drawn to different task-relevant locations. Based on this, they suggested that attention first chooses objects of interest and then 
fixations are drawn to the most informative points. 
This result supports previous findings on the influence of task on attention. Eyes extract visual information in a goal-oriented anticipatory fashion even when single actions are to be performed on the same object.


Inspired by the salient object detection models in computer vision (i.e., defining saliency at the level of objects as in~\cite{liu2011learning}), Dziemianko et al.~\cite{dziemiankoobject} applied models of salient object detection to fixation prediction, similar to Borji et al.~\cite{borji2013stands}. They 
implemented and evaluated three models of salient object detection on fixations over two tasks\footnote{This data is available at: http://homepages.inf.ed.ac.uk/keller/resources/.}:
1) visual counting: counting the number of occurrences of a cued
target object and 2) object naming: naming objects
present in the scene. In their analysis, they inserted a Gaussian blob at the center of a bounding box around an object. They showed that the object-based interpretation
of saliency provided by these models is a substantially better
predictor of fixation locations than traditional pixel-based
saliency. This result is in alignment with findings by Borji et al.~\cite{borji2013stands}.

Xu et al.\cite{xu2014predicting} studied the effects of several types of attributes on gaze guidance during free-viewing at three levels: the \textit{pixel-level}, the \textit{object-level}, and the \textit{semantic-level}. Pixel-level attributes included contrast, edge content, color, etc. Object-level attributes included size, convexity, solidity, complexity, and eccentricity. Semantic high-level attributes contained smell, sound, face, text, taste, touch, watchability, and operability. Using images with annotated objects and regression, they learned which factors were important in predicting fixations (e.g., faces and text were more important, but sound and motion less so). One of the factors they considered (categorized under object- or semantic-level attributes) was object center-bias.  
They fitted a two-dimensional normal distribution to the spatial distribution of the fixations in the object-centered coordinate system and used it to weight the object center\footnote{They did not specifically mention how they defined the center of an object or whether they used bounding boxes. It seems, however, that similar to Nuthmann \& Henderson~\cite{nuthmann2010object} and Dziemianko et al.~\cite{dziemiankoobject} they used bounding boxes.}. Although they found that adding object- and semantic-level attributes increased fixation prediction performance, unfortunately they did not explicitly measure the `added value' of object-center bias.


Several works have used object information to build attention models at the object level (e.g.,~\cite{yun2013exploring,sun2008computer,chang2011fusing,m2013fixations,kavak2013visual,Elazary_Itti08jov,yanulevskaya2013proto}).
Some of these models propose how attention should be deployed to different objects at different times to fulfill a task. Some others, similar to our goal here, have explained fixations in the context of free-viewing.
For example, Kavak et al.~\cite{kavak2013visual} used a bank of object detectors to give higher weight to regions inside objects. Recently, Stoll et al.~\cite{stoll2015overt} also proposed an approach to account for object driven fixations and concluded that objects predict fixations better than saliency when combined with bottom-up saliency. 


Despite some previous evidence for the object-center hypothesis, three challenges still exist that need to be resolved. \textit{First:} the fact that observers tend to look near the center of objects could be because saliency might also be high in those regions. In other words, do observers look at the center of object simply because saliency is higher there compared to at the object boundary? 
Nuthman et al. did not directly control for this confounding factor. Instead, they measured the distribution of saliency at salient patches/proto-objects and showed that compared to the distinct PVL for real objects, there was less evidence for a PVL for human fixations within saliency proto-objects. But this analysis does not seem to address this confound. Instead, here we measure the magnitude of low-level saliency inside the object. In a complementary analysis, we combine both saliency and object center-bias to see whether or not there is added value.

\textit{Second:} how we can define the center of an object? This is a challenging task due to variety of object parameters such as shape, size, concavity/convexity, symmetry, etc. Almost all previous studies have used bounding boxes which might not be a good option in many cases (e.g., the center of the bounding box may fall outside of the object area for a concave object). Further, using bounding boxes causes confusion and inaccuracy in assigning fixations to the foreground object or background. For example, in the analysis of Nuthman et al. in Figure~\ref{fig:Nuthman}.b, several points from the background are also included. To address this challenge, we first use object boundary polygons instead of bounding boxes. Second, we apply object center-bias on each individual object from its center of mass\footnote{The center of mass (CoM) is calculated using the standard methods. The $x$ and $y$ coordinates of the CoM are, respectively, the average of the $x$ coordinates of the pixels and the average of the $y$ coordinates of the pixels that make up the object.} towards the outside. 

\textit{Third:} this challenge is in regards to the complexity of stimulus set, since natural scenes are inherently complex. For example, observers may have different viewing behavior depending on the complexity of the scene. They may visit the center of the object for an image with few (large) objects but may not do so for objects amidst scene clutter. In order to answer this question, one needs large amounts of data. To address the challenge of complexity, we run our experiment over a large amount of data from two datasets with a variety of images and objects.

\subsection{Contributions}

In summary, we offer the following contributions in this work:
\begin{enumerate}
\item We verify the hypothesis that ``observers tend to look near the center of objects in scene free-viewing'' and establish that this effect is independent of low-level bottom-up saliency. 
\item We construct a combined model of object center-bias and saliency. To do so, we answer the following questions: a) How can we construct an object center-bias map to emphasize object centers? b) What is the best way to combine this map with image saliency (addition or multiplication)?

\end{enumerate}

\section{Data}

\subsection{Our Data}

\subsubsection{Stimuli}
\begin{figure*}[t]
\centering	
\includegraphics[width=\linewidth]{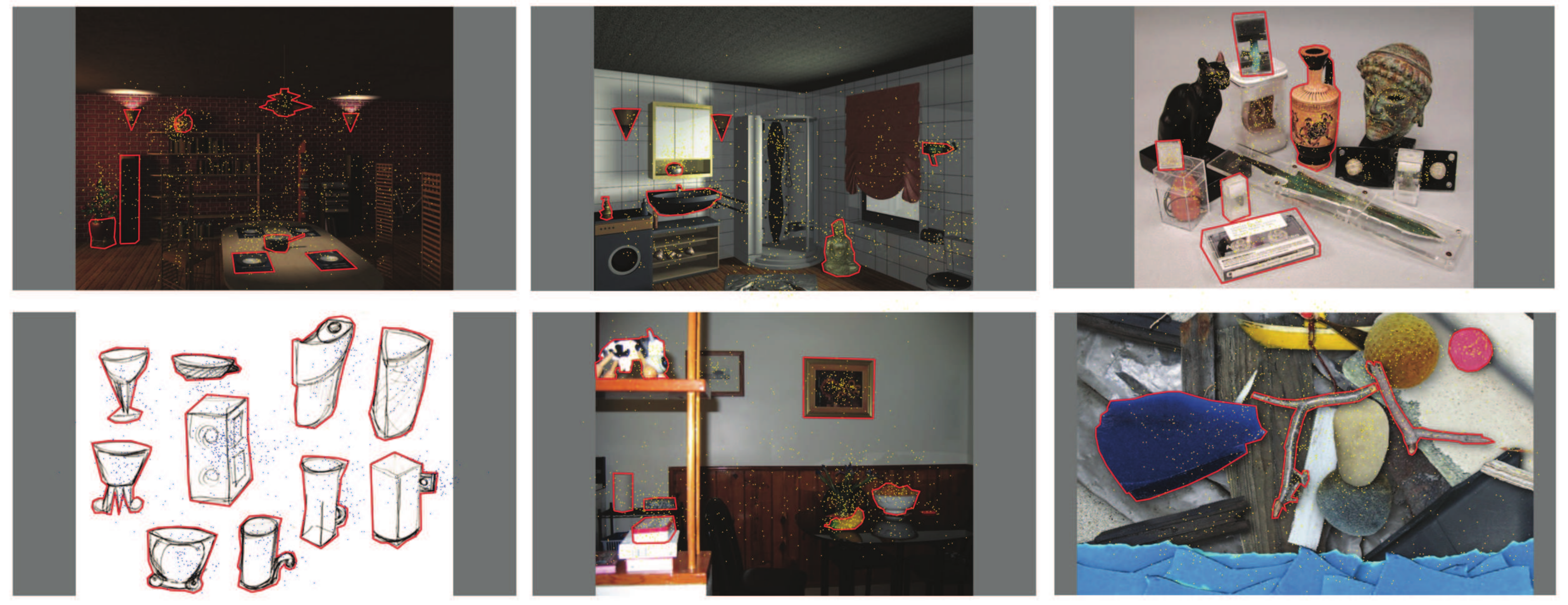} 
\caption{Sample images from our dataset along with annotated objects and fixations of all observers. 
Notice how certain locations inside some objects attract more fixations than others.}
\label{fig:ourData}
\end{figure*}

Stimuli consisted of 60 color images (30 synthetic, 30 natural). Figure~\ref{fig:ourData} shows some examples of our stimuli. Images were resized to 1920 $\times$ 1080 pixels by adding gray margins while preserving the aspect ratio. We intentionally did not include stimuli with persons, animals, or faces, mainly because 
these objects have interesting parts on their ends. We chose images from different categories (line drawings, 3-D rendered cartoonic images, etc.) with different types of objects. Object boundaries were manually traced. 
Our methodology for selecting objects was to only label objects that were completely unoccluded in the image. This was done so that the analysis of a center bias effect would not be influenced by objects whose computed center of mass was different from the theoretical center of mass. We attempted to choose images with less photographer bias\footnote{Tendency of photographers to frame interesting objects at the center of the image.} and with multiple objects off the image center, thus reducing the effect of center-bias on fixations.


\subsubsection{Observers}
Seventeen observers (4 male, 13 female) participated in this experiment (mean age = $20.58$, std = $1.37$).
Observers were students at the University of Southern California (USC) from the following majors: 
Neuroscience, Psychology, Biology, Business, Biomedical Engineering, and Accounting.
The experimental methods were approved by the USC's Institutional Review Board (IRB). Observers had normal or corrected-to-normal vision and were compensated by course credits. Observers were asked to freely watch the images.

\subsubsection{Apparatus and procedure}
Observers sat 130 cm away from a 42 inch monitor screen such that scenes subtended approximately $43^{\circ} \times 25^{\circ}$ of visual angle. A chin/head rest was used to minimize head movements. 
Stimuli were presented at 60Hz at a resolution of $1920 \times 1080$ pixels in random order. Eye movements were recorded via an SR Research Eyelink eye tracker (spatial resolution of 0.5$^{\circ}$) sampling at 1000 Hz. Each image was shown for 30 seconds followed by a 5 seconds delay (gray screen). The eye tracker was calibrated using a 5-point calibration method at the beginning of each recording session.

\subsection{OSIE dataset}

\begin{figure*}[t]
\centering	
\includegraphics[width=\linewidth]{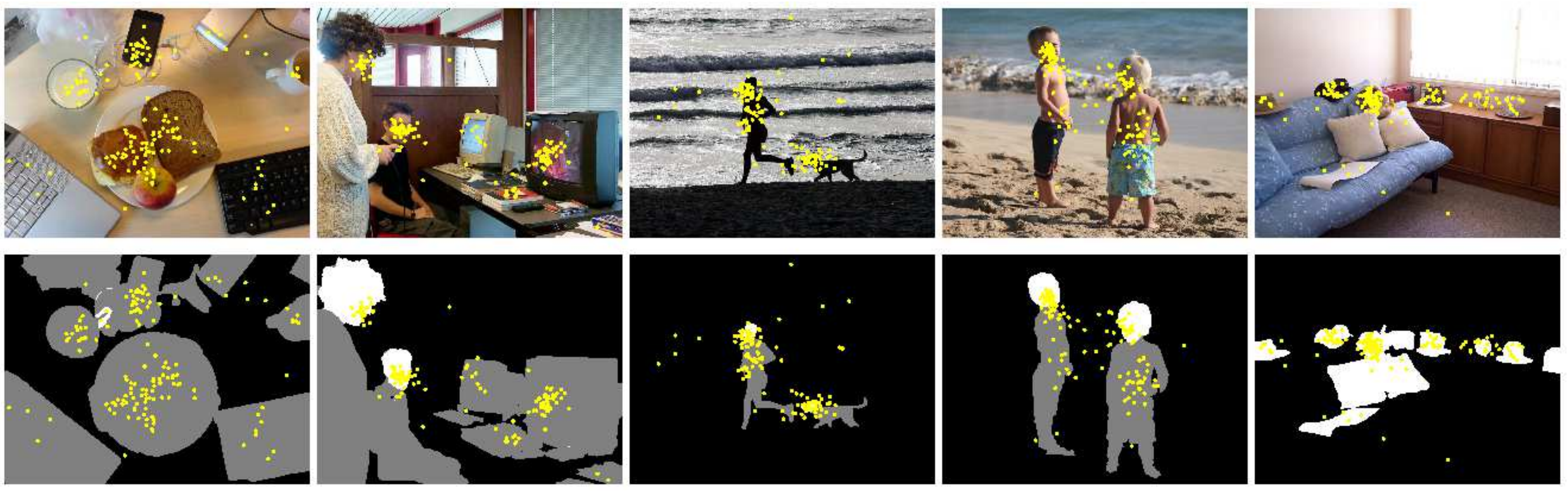} 
\caption{Sample images from the OSIE dataset along with object annotations and fixations. Due to shorter presentation times (5 seconds vs. 30 seconds in ours), there are fewer fixations in OSIE images than in ours.}
\label{fig:OSIE}
\end{figure*}

The OSIE (``Object and Semantic Images and Eye-tracking'') dataset\footnote{http://www.ece.nus.edu.sg/stfpage/eleqiz/predicting.html} was created by Xu et al.,~\cite{xu2014predicting} to explore how object and semantic saliency can be used for predicting where observers look in free viewing of natural scenes. It contains eye tracking data of 15 participants over a set of 700 images (for 3 seconds viewing time).
Each image has been manually segmented into a collection of objects by one person. Semantic attributes of objects have also been manually labeled (e.g., operability, watchability, text). This dataset introduced
two novel contributions: First, it contains a large number of object categories and several objects have semantic meanings and second, the majority of the images contain multiple dominant objects.
Figure~\ref{fig:OSIE} shows example images from the OSIE dataset along with fixations and object annotations. Please refer to Xu et al.~\cite{xu2014predicting} for more details on this dataset.

OSIE dataset is suitable for our purposes because it has a variety of images from different categories. Further, object boundaries have been carefully annotated on this dataset for a large number of objects.

\begin{figure}[t]
\centering	
\includegraphics[width=\linewidth]{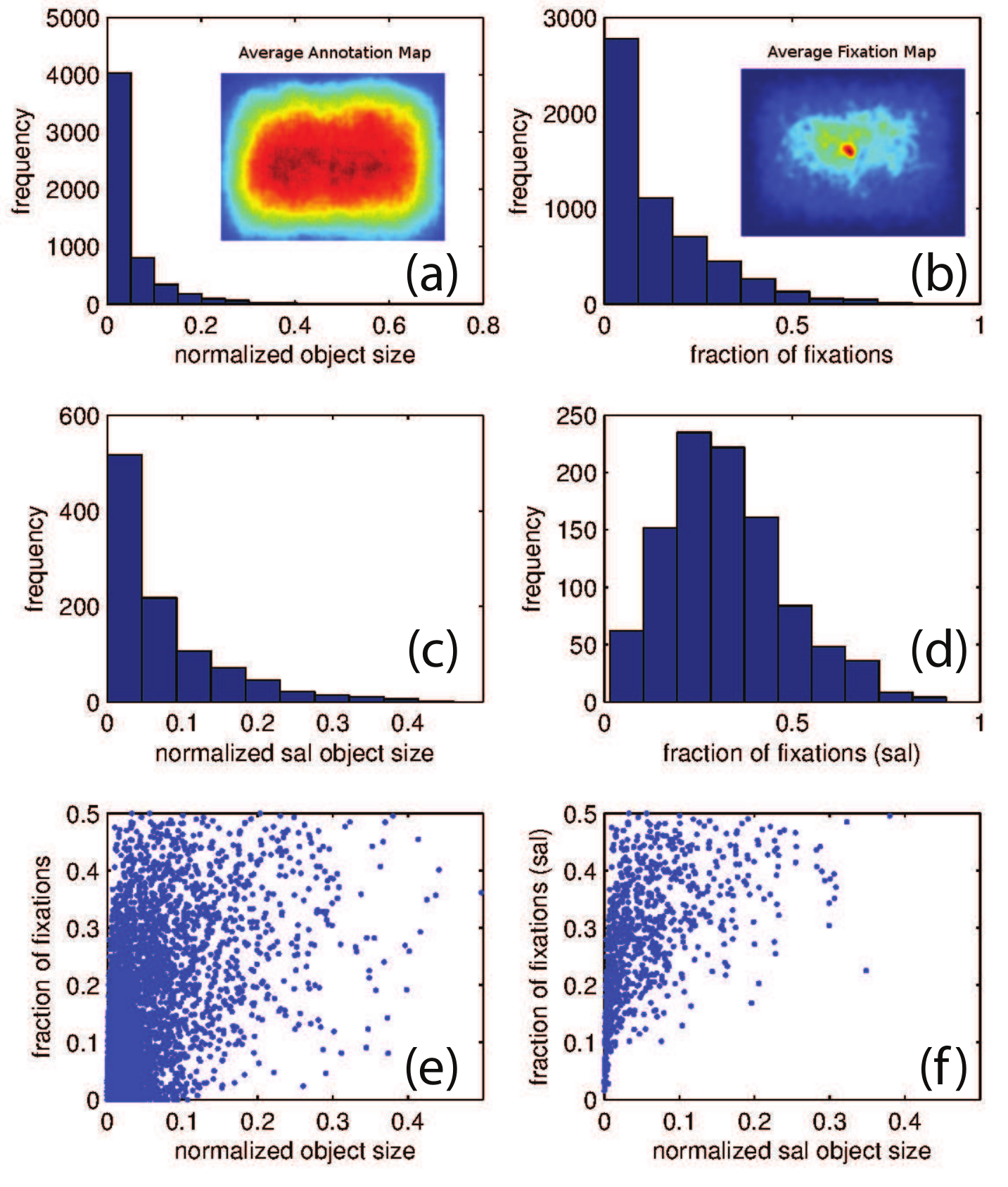} 
\caption{Statistics of the OSIE dataset. a) histogram of normalized object size, b) histogram of fraction of fixations (number of fixations on an object over number of all image fixations), c) histogram of normalized salient object size. Salient object is the one with the maximum fraction of fixations on it, d) similar to b but for salient objects, e \& f) plot of fraction of fixations as a function of normalized object size. `Frequency' on the y-axis indicates the number of occurrences.
 }
\label{fig:statsOSIE}
\end{figure}

Figure~\ref{fig:statsOSIE} illustrates statistics of the OSIE dataset. The majority (87.01\%) of objects occupy equal or less than 10\% of the image area.
52.68\% of objects contain equal or less than 10\% of the fixations on the image. We observe that 
normalized size of the most salient object (object at the peak of the fixation map; 1012 out of overall 5551 object annotations) is usually larger than regular objects as shown in 
Figure~\ref{fig:statsOSIE} second row. 74.90\% of most salient objects occupy equal or less than 10\% of the image area. Similarly, only about 5\% of most salient objects contain equal or less than 10\% of the fixations on the image. About 14\% of the most salient objects contain equal to or more than 50\% of the fixations in the image. This can also be observed from the third row of Figure~\ref{fig:statsOSIE} which shows the relationship of normalized object size versus the fraction of fixations over all object annotations. Insets in Figure~\ref{fig:statsOSIE} show the average annotation map and average fixation map. As in other eye movement datasets, a large degree of fixation center-bias
is observed on this dataset.

\begin{figure*}[t]
\centering	
\includegraphics[width=.9\linewidth]{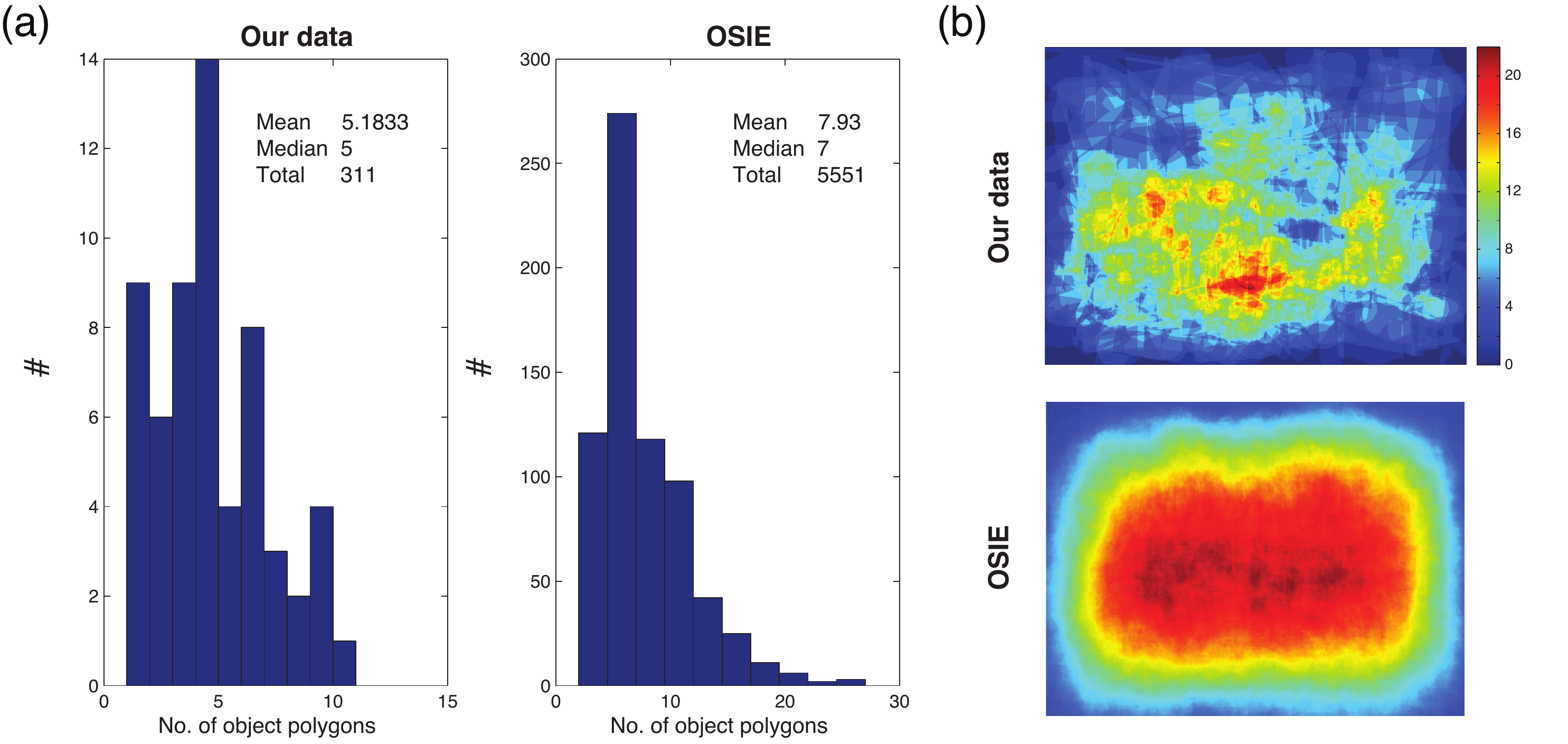} 
\caption{(a) Histograms of annotated objects per image over the two datasets. Images in the OSIE dataset contain more object annotations on average compared to our dataset. (b) Average object annotation map over two datasets.}
\label{fig:HistObj}
\end{figure*}

On average, 5.18 and 7.93 objects are annotated over our dataset and OSIE, respectively (median: 5 vs. 7).
The total number of fixations on our dataset is 76,869 (over 60 images). This figure for OSIE dataset is 98,321 (over 700 images). Figure~\ref{fig:HistObj} shows a histogram of annotated objects and the average annotation map over the two datasets.

\section{Measuring object center-bias}

\begin{figure*}[!htbp]
\centering	
\includegraphics[width=0.8\linewidth]{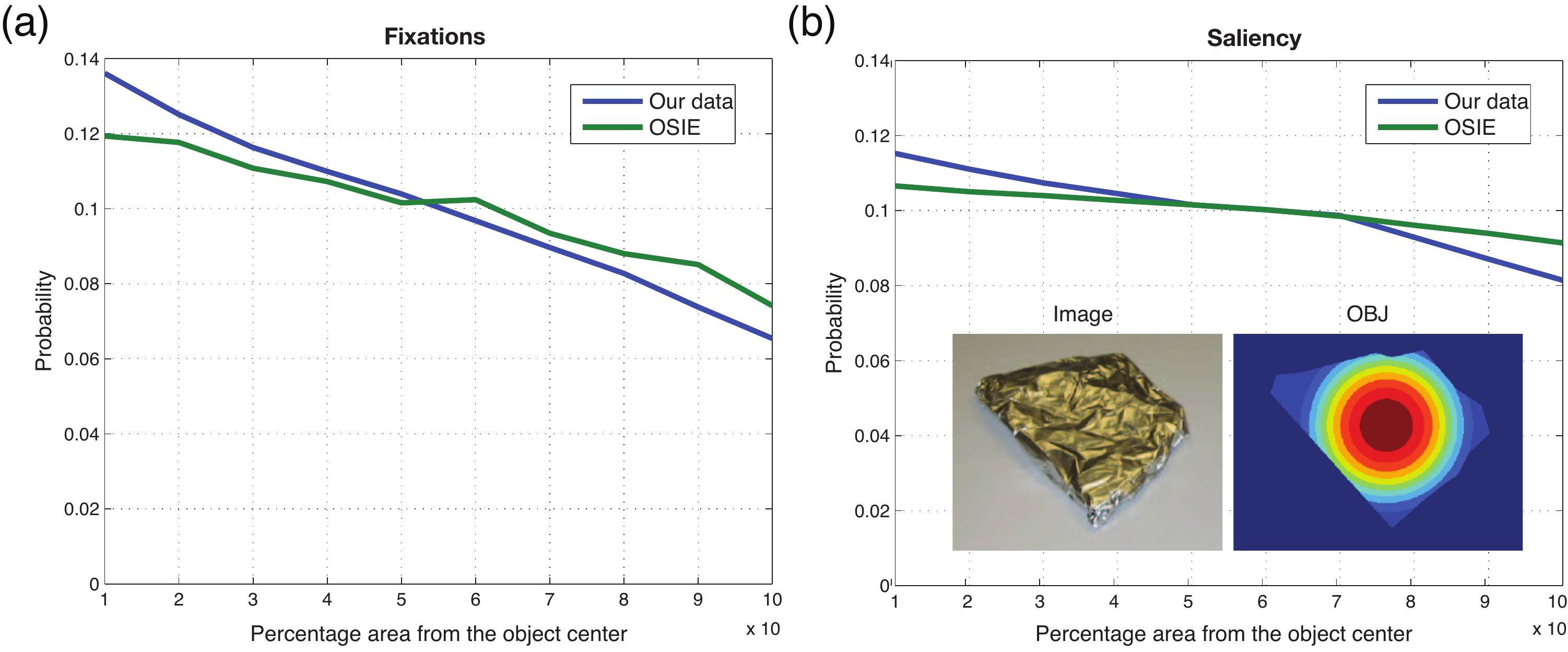} 
\caption{(a) Distribution of fixations over the object area from the inner-most ring (1 in x-axis) to the outer-most ring (10 in x-axis). Note that the difference in rings adds $10\%$ to the object area and not the entire circle (i.e., it is incremental). (b) Distribution of saliency using the AWS saliency model, over both datasets. Inset shows an example object and the corresponding object map (denoted OBJ).}
\label{fig:hist}
\end{figure*}

In this section, we verify the object center-bias hypothesis by measuring the distribution of fixations inside objects. To do so, we need a way to define the center of an object. We choose the center of mass of an object as the object center. Then, we grow circles from the object center such that each circle (tube) contains an additional $10\%$ of the object area. In other words, the difference of object coverage between each successive pair of concentric circles is $10\%$ of the whole object area. We repeat this operation until all object area is covered, Figure~\ref{fig:hist}.b inset shows an example of this operation. We call this map, "object center-bias map" and denote it by "O".

For each of the circular regions (tubes), we then count the number of fixations that fall on that region. Figure~\ref{fig:hist}.a shows the distribution (converted to probability density function) of fixations over the 10 circles averaged over all objects on each dataset. As it shows, as one moves away from the object center toward the object boundary, the probability of fixations declines (almost linearly). 

Figure~\ref{fig:hist}.b shows the 
distribution of saliency (average saliency inside each tube) using the AWS saliency model~\cite{GarciaDiazJOV} from center to boundary of the objects. Here, again we observe a decline in saliency as moving from object center toward the object boundary. Similar to fixations, this decline is sharper on our dataset than on the OSIE. This result indicates that on average, saliency is higher at the object center which, as discussed in the introduction, may explain some of the additional fixations in that region. To answer whether saliency can explain all fixations or not (i.e., discounting the effect of saliency confound), in the next section we follow a modeling approach by adding these two components. The rational is as follows: if we observe a boost in saliency in predicting fixations by adding object center-bias, we can then conclude that object center-bias has an (independent) added value to what early saliency already offers.

\begin{figure*}[t]
\centering	
\includegraphics[width=.9\linewidth]{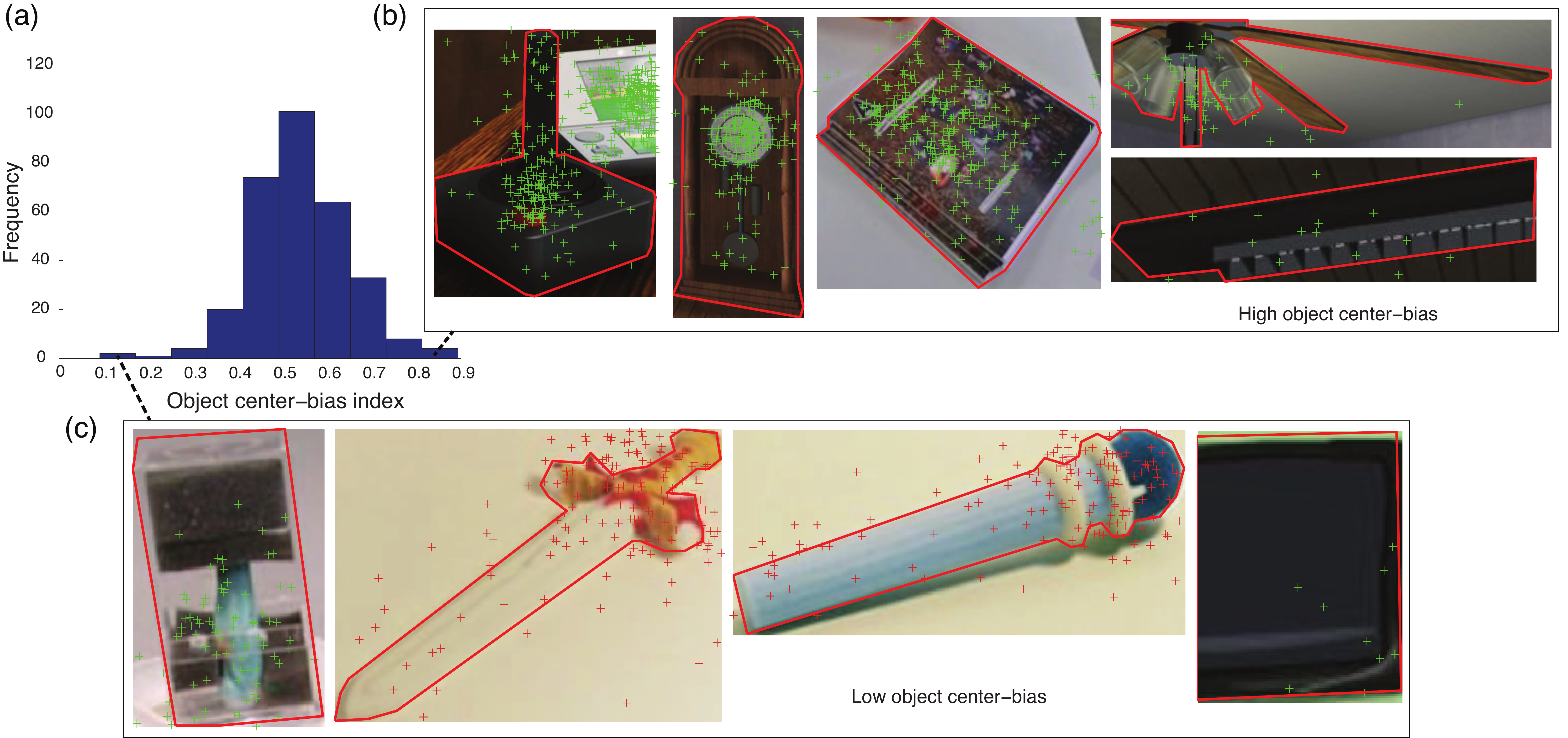} 
\caption{(a) Histogram of object center-bias indices over our data. An index above 0.5 means more center-bias. (b) Some objects with high indices, (c) Some objects with low indices. These objects usually have a salient part on one of their ends.}
\label{fig:centerBiasHist}
\end{figure*}

To explore the generality of the hypothesis over all objects and the factors that it may depend on, we define an object center-biased index which is the sum of fixation densities inside the first 5
inner-most circles/rings over the sum of fixation densities inside all ten circles/rings (i.e., over the entire object):
\begin{equation}
obj\_cnt\_idx = \frac{\sum_{i=1}^5 p_i}{\sum_{i=1}^{10} p_i}
\end{equation}
where $p_i$ is the density of fixations inside the i-th tube. The higher the obj$\_$cnt$\_$idx, the more tendency of fixations towards the object center. Figure.~\ref{fig:centerBiasHist} demonstrates the histogram of obj$\_$cnt$\_$idx indices on our dataset. For the majority of objects (200 out of 311) this index is higher than 0.5, which would be the value if fixations were distributed uniformly over the entire object. As expected, objects with high obj$\_$cnt$\_$idx often have content at the image center (Figure.~\ref{fig:centerBiasHist}.b, e.g., book, grandfather clock) while objects with low obj$\_$cnt$\_$idx usually have imbalanced/tilted features on one side (Figure.~\ref{fig:centerBiasHist}.c, e.g., sword, microphone). 
We notice that affordance and shape of the object also influences where people look inside it. For example, in the microphone case, there are more features around its tip including salient edges which differ from their neighbors (hence high saliency there) which attracts more fixations (similar argument for the sword). Replacing the circles with bounding boxes (i.e., rectangular tubes) shows the same pattern of results.


\section{Our augmented saliency model}


Having seen that object center-bias effect exists on a majority of objects, in this section we propose a simple combined model of saliency and object center-bias. This model, in addition to having better fixation prediction accuracy, also helps further investigate the accuracy of the object center-bias hypothesis. We follow the previous line of research that linearly combines cues for computing saliency (e.g.,~\cite{2009Cerf,judd2009learning}). Our model is simply a weighted combination of the saliency map and the object center-bias map as follows:

\begin{equation}
 SM = (1 - \beta) \times S + \beta \times O, \ \ \beta = 0:0.1:1
\end{equation}

\noindent where S is the saliency map, O is the object center-bias map, and $\beta$ is a parameter that controls the relative magnitude of the two maps. $\beta=0$ is just the pure bottom-up saliency map (AWS model), and $\beta=1$ is the pure object center-bias map. 
Through experiments, we learned that adding the term $S \times O$ did not improve our results, so we discard it here. The S, O, and resulting SM maps are all normalized (sum to 1).

\begin{figure*}[t]
\centering	
\includegraphics[width=.8\linewidth]{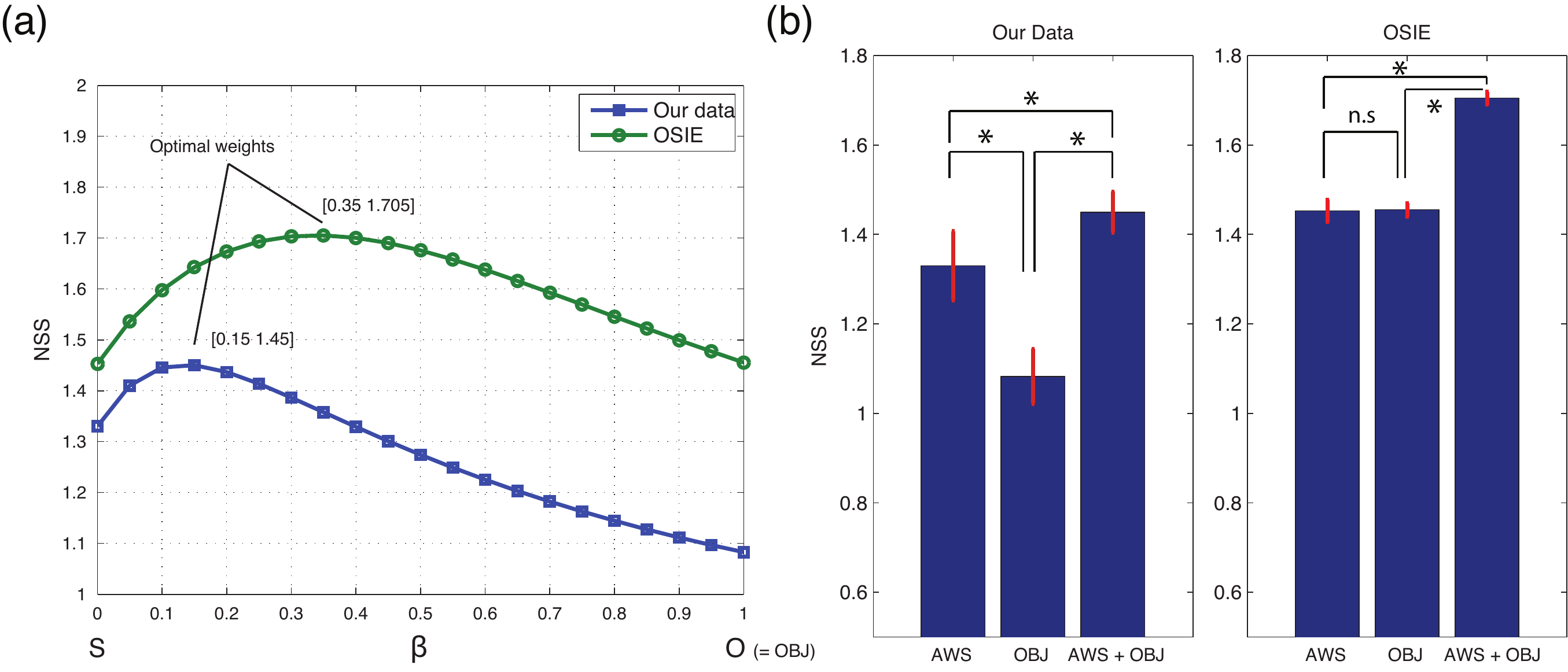} 
\caption{(a) NSS score of the combined model as a function of $\beta$. $\beta = 0$ corresponds to pure saliency and $\beta = 1$ corresponds to the pure object map. Note that over the whole range of $\beta$ values, the combined model performs better than both the saliency and object models over the OSIE dataset. Over our data, since some objects are annotated and not all, a larger magnitude of the saliency model is necessary to make a superior combined model. (b) Average NSS score of the models over our data and the OSIE dataset. Error bars indicate standard error of the mean (s.e.m).}
\label{fig:barRes}
\end{figure*}

Figure~\ref{fig:barRes}.a shows the NSS\footnote{Normalized Scanpath Saliency~\cite{Peters_etal05vr}, which is the average of the response values at human eye positions in a model's saliency map that has been normalized to have zero mean and unit standard deviation. NSS = 1 indicates that the subjects' eye positions fall in a region whose predicted saliency is one standard deviation above average. NSS $\leq$ 0 indicates that the model performs no better than picking a random position, and hence is at chance in predicting human gaze.} scores of the combined model as a function of parameter $\beta$. As $\beta$ increases, the NSS peaks and then declines over both datasets. Looking at the optimal $\beta$ for each dataset, we find that they are close to each other, 0.15 for our data and 0.35 for OSIE, which result in NSS scores of 1.45 and 1.705, respectively. 
This means that if we were to train the model over our data and test it on the OSIE dataset (or vice-versa), we would have achieved a better performance than both saliency and object center-bias maps on the destination test dataset. In other words, if we were to apply the best $\beta$ from one dataset to another, results would be still better than both saliency and object center-bias models. This means that our model generalizes well over datasets.

Figure~\ref{fig:barRes} also shows higher performance over OSIE dataset compared to our dataset which can be attributed to two causes: 1) more objects are annotated in OSIE images than our images which results in a higher contribution of objects (mean 5.18 on our data vs. 7.93 over OSIE), and 2) viewing time is longer on our data which might have caused subjects to be driven more by the image background. We believe that the second cause is a more plausible explanation of this effect as we did not see a trend in performance as a function of the number of annotated objects on a scene. Further, while the number of images over OSIE dataset is about 12 times higher than our data, the number of fixations is nearly the same. Longer viewing time leads to fixations that fall on the background clutter and this results lower prediction accuracy since these fixations are not accounted by the object annotations.

Figure~\ref{fig:barRes}.b shows the results over both datasets for saliency alone, object map alone, and their optimal combination. Average NSS for AWS, OBJ (i.e., object center-bias map O), and the combined model (with optimal $\beta$) over our data in order are: 1.3302, 1.0828, and 1.4501. Combined model significantly outperforms the other two models (t-test, combined vs. AWS, p = 1.9301e-06; combined vs. OBJ, t-test, p = 2.9015e-16). AWS model here significantly outperforms the OBJ model (t-test; p = 7.0320e-06). 

The average NSS for AWS, OBJ, and combined model over OSIE dataset in order are: 1.4530, 1.4554, and 1.7051.
The combined model significantly outperforms the other two models (t-test, combined vs. AWS, p =3.1412e-69; combined vs. OBJ, t-test, p = 1.9295e-73). The difference between AWS and OBJ models is not statistically significant here (t-test; p = 0.9136). The difference between the combined model and the saliency model is smaller in our dataset compared to the OSIE dataset (9\% vs. 17.27\%). This could be due to the larger number of annotated objects in the OSIE images than in the images in our dataset. Interestingly, on OSIE, all tested values of $\beta$ other than 0 and 1 are above both AWS and the object center-bias models. Our object center-bias model is essentially similar to the model proposed by Einh{\"a}user et al.~\cite{Einhauser2008objects} with the difference that here we emphasize the object center instead of uniformly distributing activity over the entire object. Further, there is no object weighting based on memory recall (i.e., the same weight for all objects).

\begin{figure*}[!htbp]
\centering	
\includegraphics[width=\linewidth]{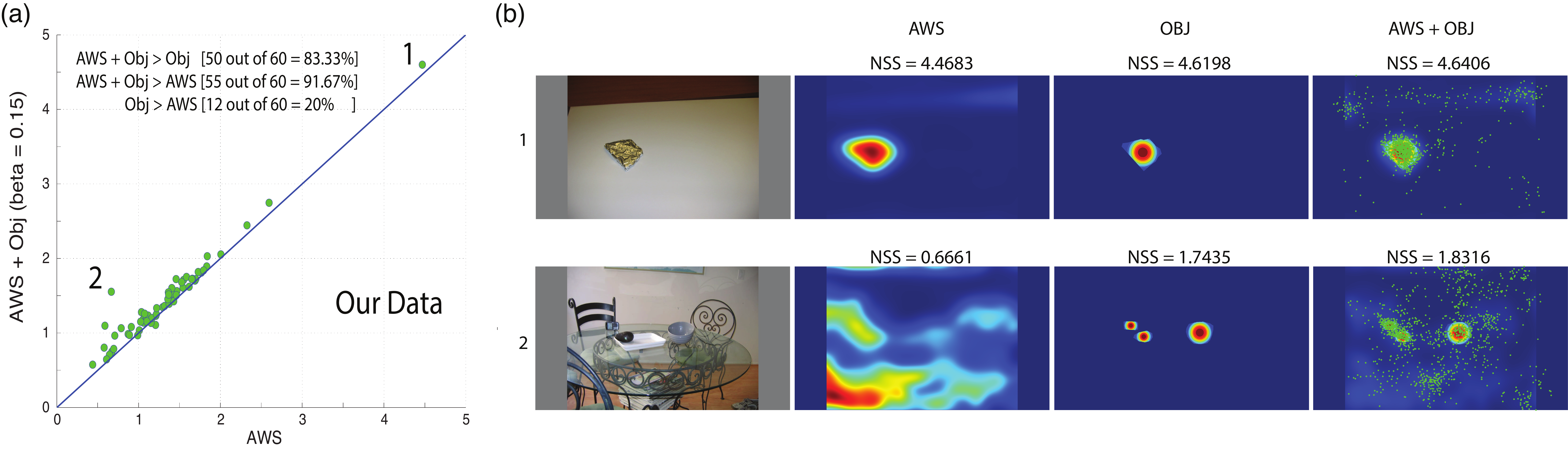} 
\caption{(a) Image-wise comparison between the NSS score of saliency vs. a combined model for all images in our data. Each dot is for one image. Performance of the combined model is with the optimal $\beta$. Even with a small number of annotated objects per image, we observe an increase in performance of the combined model. 
The percentage of images for which the combined model performs better than each individual component is also shown. For 55 images, the combined model outperforms the AWS model (50 with respect to the OBJ model). (b) Two images with their corresponding prediction maps. For the first image, the saliency map already explains many of the fixations (i.e, high NSS) so inserting object center-bias, although helpful, does not add much to the score. For the second image, the object map brings a lot of value.}
\label{fig:scatterOur}
\end{figure*}

\begin{figure*}[!htbp]
\centering	
\includegraphics[width=\linewidth]{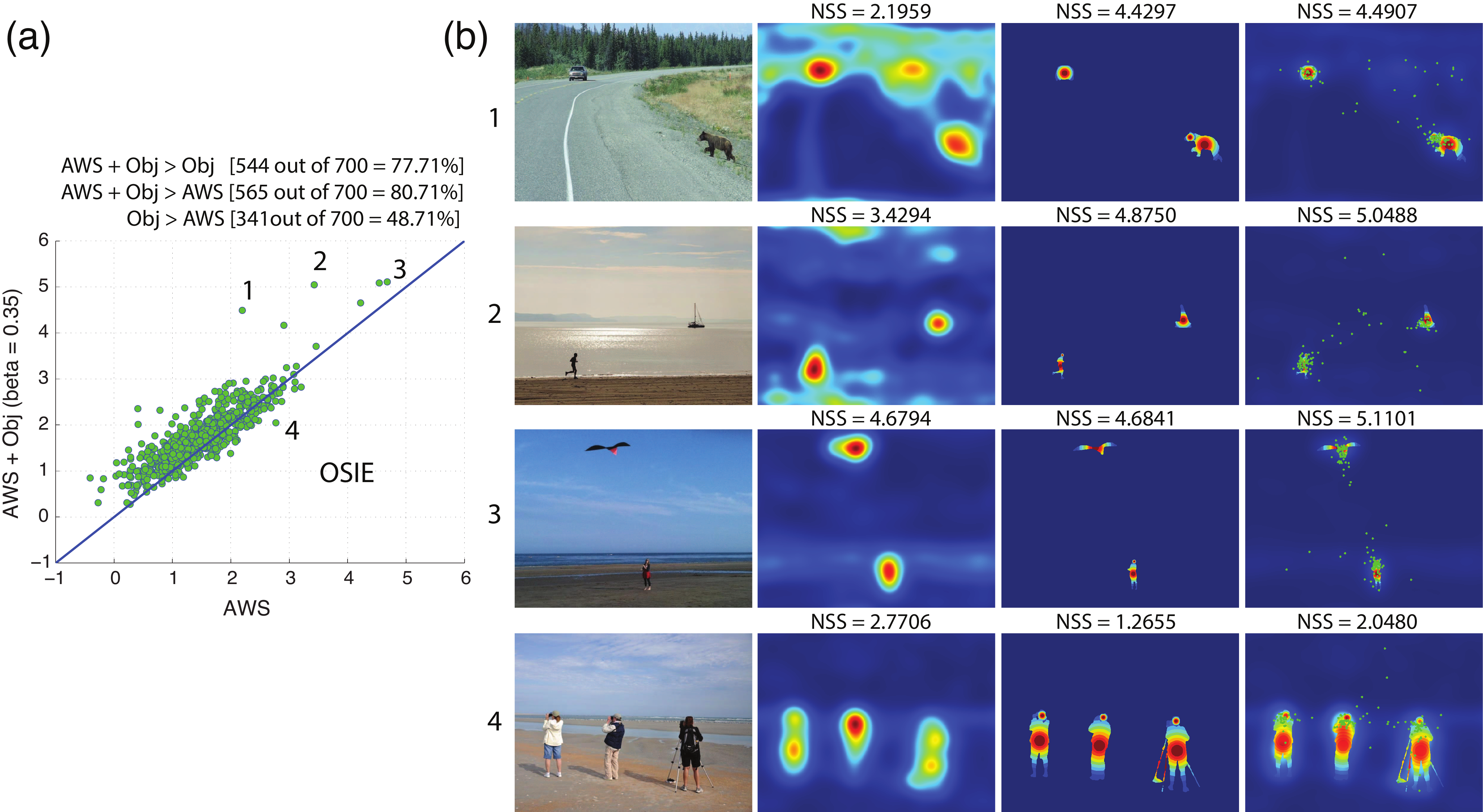} 
\caption{Similar to Figure~\ref{fig:scatterOur} but over the OSIE dataset. (a) NSS score of saliency vs. combined model, b) Sample images with their corresponding prediction maps. These images were chosen to show cases where map combination increases performance (compared to AWS) drastically (images 1 \& 2), moderately (3), and a case where combination slightly hinders performance (4). On image 4, each person was annotated as one object and emphasis was placed at the center of their body while fixations were drawn to their heads. Better performance would have been achieved if human heads were annotated on this image.}
\label{fig:scatterOSIE}
\end{figure*}

Figures~\ref{fig:scatterOur} and~\ref{fig:scatterOSIE} show scatter plots of saliency vs. combined model over our data and OSIE, respectively. Each dot in this plot represents the NSS score for one image. Over our dataset, for 91.67\% of images, the combined model outperforms the AWS saliency model. This figure for the OSIE is 80.71\%. These values for the combined model vs. object center-bias map over our data and OSIE, in order are 83.33\% and 77.71\%. On both datasets for less than 50\% of the images, the object map wins over the saliency map (20\% on our data and 48.71\% over OSIE). For images where the combined model outperforms the saliency model significantly, there are usually few objects in the scene (e.g., Figure~\ref{fig:scatterOSIE}.b, images 1, 2, and 3) and scenes do not usually have much background clutter. For images where the combined model performs worse than saliency, usually interesting parts of the object do not happen at the object center (e.g., in people, where the entire body is annotated as one object, face is the most interesting part but it is not at the center; Figure~\ref{fig:scatterOSIE}.b, 4th image).

\begin{figure*}[t]
\centering	
\includegraphics[width=.8\linewidth]{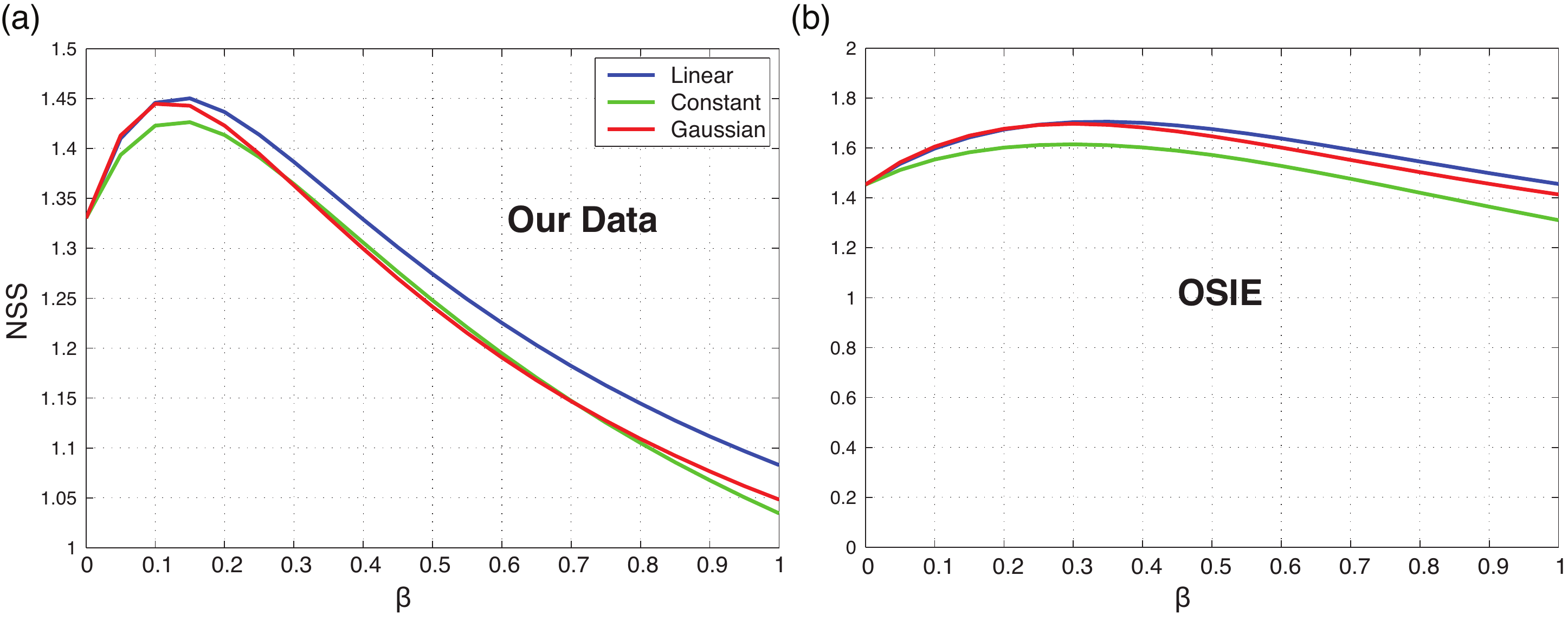} 
\caption{NSS score with different types of object center-bias emphasis over (a) our data, (b) OSIE dataset. Results do not show a big difference in performance or in optimal $\beta$. It seems that linear weighting is the best strategy over both datasets.}
\label{fig:centerBias}
\end{figure*}

Figure~\ref{fig:centerBias} shows the NSS score for three different 
types of object center-bias including linear weighting (our implementation so far), constant weighting (uniform distribution of weight over the entire object), and Gaussian weighting (which weights the 10 circles/rings using a normalized Gaussian function) over (a) our data and (b) OSIE dataset. Results do not show a big difference in performance or in optimal $\beta$. We find that linear weighting of object center-bias is the best strategy consistently over both datasets.

We also noticed that replacing polygons with bounding boxes (similar to~\cite{nuthmann2010object}) over OSIE dataset results in NSS of 1.112 which is above NSS of 1.083 using polygons but overall does not significantly improve the combination performance. The higher performance using bounding box is because it better accounts for fixations around the edges of objects.



\section{Discussion}

In this work, we verified the validity of the object center-bias hypothesis in the context of free-viewing. We believe there might be an even stronger effect of object center-bias in the presence of a task. According to the \textit{cognitive relevance theory} (see~\cite{hayhoe2003eye}) objects are more important when there is a task (compared to free-viewing). Some interesting tasks in this regard include: 
1) Asking subjects to count the number of objects in a scene, 2) Asking subjects to manipulate objects (e.g., in a coffee-making task). In the latter, subjects may also look at those features that are
related to the task (e.g., handle of the kettle) as suggested in Belardinelli \& Butz~\cite{belardinelligaze}. It has also been shown that in object categorization, human subjects fixate on informative parts of objects (See Hartendorp et al.~\cite{hartendorp2013relation}). Some other interesting tasks here include: aesthetic judgment, interestingness judgment, visual search, and scene memorization.

Here, we discuss some important parameters for further investigation of the object-center hypothesis that should be taken into account in future studies. The first parameter is scene clutter. The manner in which humans attend to objects might be different depending upon whether they are viewing a simple scene with few objects or a complex scene with several objects and/or an amorphous background. In a complex scene, viewers may quickly scan the image in order to collect more information which may cause them to be driven to spatial outliers. The second parameter, related to the first one, is scale.
If objects are shown to observers in a large scale (and hence larger objects sizes), then they may not tend to look at the empty central regions inside the object specially if they don't contain features (imagine close up view of a white board).
The third parameter concerns object symmetry. It has been shown in Kootstra et al.~\cite{kootstra2011predicting} that people tend to look at the center of symmetrical objects. The question that arises here is ``Are object center-bias and symmetry two different cues?''. In other words, ``Do people look at the center of asymmetrical objects?''. The fourth parameter regards viewing constellated objects made of several components. Object concavity/convexity is the fifth parameter. For example, what happens if the center of the object lies outsides the object? 

To investigate above-mentioned parameters we recommend two approaches: First, more systematic studies over simple synthetic scenes are desirable. For example, imagine a plain object with no features inside. As soon as a salient point/region is inserted somewhere inside the object (but off-center), most likely viewers will not look at the center anymore (or will look less). This is in alignment with our analysis in this paper which was testing whether saliency peaks at the center of the objects in the real world or not. Another similar analysis would be collecting objects with no salient points inside and test whether viewers still look around the object center (similar to some of our images). Overall, the main difficulty in investigating the object center-bias arises from the fact that there is large variety of objects in natural scenes. Indeed, the object-center effect is stronger for some certain types of objects. Second, we believe 
that large scale object annotated datasets (e.g., datasets by Greene~\cite{greene2013statistics}\footnote{http://stanford.edu/~mrgreene/labelme.html}, Cheng et al.,~\cite{cheng2014salientshape}\footnote{http://mmcheng.net/gsal/}, and Li et al.,~\cite{li2014secrets}\footnote{http://cbi.gatech.edu/salobj/}) can be very useful to understand how saliency and object information are related in scene viewing and understanding. 

In contrast to Nuthmann \& Henderson's conclusion~\cite{nuthmann2010object} which stated that ``... attentional selection in scenes is object-based. Saliency only has an indirect effect on attention, acting through its correlation with objects ...'', our results suggest that both low-level saliency and object information (here object center-bias) contribute (although correlated) to attention during scene free-viewing. This finding aligns with our previous results in Borji et al.~\cite{borji2013objects} where we criticized the hypothesis by Einh{\"a}user et al.~\cite{Einhauser2008objects} that
``Objects predict fixations better than early saliency'' and showed that saliency is a better predictor of fixations in free-viewing\footnote{At least with the way that Einh{\"a}user et al. used objects to build a model. If they had added object-center bias to their model, most likely they would have achieved much better results compared with saliency alone (i.e., the OBJ model in our work).}. Einh{\"a}user et al., built a map with object regions weighted by their recall frequency in a scene viewing (for memory testing) task. Although the debate whether saliency or objects are better predictors of fixations is still ongoing, the bottom-line is that both factors contribute independently to guiding fixations.

Is object center-bias a bottom-up or top-down cue? It is true that object center can be computed by a simple computationally-efficient early processing (using proto-objects~\cite{rensink2000dynamic}) but the mechanism that chooses to drive saccades to the center of objects (even in presence of more salient edge regions) seems to be a top-down process. By analogy to the face cue that attracts attention and gaze, there might be some dedicated neural circuitries for driving saccades to the object center. This is in alignment with the object-based theory of attention which states that objects are the unit of attention. Actual implementation of this mechanism needs to be further investigated by neurophysiology and psychophysics studies.

Are eye movements driven by objects or by early saliency? And by extension, is attention object-based~\cite{duncan1984selective,egly1994shifting,rensink2000dynamic,vecera1994does,drummond2010object,gottlieb2010attention,Einhauser2008objects,henderson1993eye,nuthmann2010object,stoll2015overt} or saliency-driven~\cite{TriesmanGelade,Koch_Ullman85,1998Itti,ReinagelZador99,ParkhustEtal2002}? 
Based on our results here (as well as previous studies~\cite{xu2014predicting,stoll2015overt,dziemiankoobject,belardinelligaze,kavak2013visual,yanulevskaya2013proto}), we believe that both forms of attention guidance do occur. However, this needs to be studied further, for example by carefully controlling the scene complexity and background clutter. One approach would be using objects with no texture inside them (e.g., shapes) and see whether observers look at object centers. One piece of evidence that eye movements are driven by early saliency comes from the fact that eye movements are driven to salient regions in scenes where there are no well-defined objects (e.g., fractal scenes~\cite{Peters_etal05vr}). Evidence in favor of object-based attention comes from the finding that fixations are driven to the center of objects~\cite{nuthmann2010object,pajak2013object}. The interplay between these two forms of attention in daily life still remains to be investigated further.

Are saliency and object center-bias independent cues? In other words, do they both contribute to guiding gaze? Here, we showed that a simple linear combined map of both cues outperforms each individual map. This indirectly shows that there is an added value in their combination which means that these maps are not subsets of each other. In a more direct analysis, in a parallel study to ours, Stoll et al.~\cite{stoll2015overt} have addressed this question. They modified their stimuli by fading edges of objects (effectively reducing saliency) and then measured the performance of early saliency models versus an object center-biased model. They showed that performance of early saliency models degraded drastically over modified stimuli while performance of object center bias remained the same. From this, they concluded that saliency and object center bias are two different cues.

Some of the saliency models that have done well in previous benchmarks (e.g.,~\cite{BorjiTIP}) might have implicitly emphasized object center more (e.g.,~\cite{GarciaDiazJOV,harel2007graph}).
For example, the AWS model generates some notion of objecthood using proto-objects and whitening.
 Thus, without being fully aware of the object center-bias hypothesis, these models have been able to predict fixations better. Explicit integration of this effect into saliency models (similar to our work here) or using 
more recent models (e.g., Boosting or Conditional Random Fields (CRF)) could be an interesting direction for future modeling.

In addition to datasets used here, some other annotated datasets exist which can be used to further investigate the relationships between bottom-up saliency and object center-bias and also study the above-mentioned factors. Three examples include: 1) 
the dataset by Greene~\cite{greene2013statistics} which is mainly designed for scene categorization and understanding research (http://stanford.edu/$\sim$mrgreene/labelme.html). A total of 48,167 objects have been hand-labeled in 3,499 scenes from 16 categories using the LabelMe tool, 2) 
the UCSB dataset created by Koehler et al.~\cite{koehler2014saliency}\footnote{https://labs.psych.ucsb.edu/eckstein/miguel/research\_pages/ \\ saliencydata.html}. This dataset contains 800 images. One hundred observers performed four tasks (22 performed explicit saliency judgment, 20 performed free viewing, 20 performed saliency search, and 38 performed a cued object search task), and 3) a dataset recently introduced by Li et al.~\cite{YiHou2014} known as PASCAL-S. These authors first segment all objects and then assign saliency orders to objects. This dataset contains eye movements of 8 observers over 850 images from the PASCAL VOC dataset~\cite{everingham2010pascal}.

\section{Conclusion}


In this study, we first evaluated the object center-bias hypothesis by Henderson~\cite{henderson1993eye} and Nuthmann \& Henderson~\cite{nuthmann2010object} over two datasets in the free-viewing task. We found (results in section III) that both fixation density and bottom-up saliency are high at the center of objects, making saliency a potential confounding factor for the object-center hypothesis. To address this confound, we then proposed a combined model of saliency and object center-bias that outperforms each component significantly. This proves the object center-biased hypothesis and indicates that both saliency and object information contribute to gaze guidance in scene viewing. 
Although both saliency and object center-bias correlate with each other, neither is a subset of the other and that is why their combination performs better than each cue individually. We also noticed that this finding is consistent whether using bounding boxes or polygons, and using different saliency models or weighting approaches. Overall, our results support those of recent works that object center-bias improves fixation prediction (e.g., Xu et al.,~\cite{xu2014predicting} and Stoll et al.,~\cite{stoll2015overt}) which further support the hypothesis that fixations are driven by objects as well as early saliency.

We hope that our work will open new directions to understand strategies that humans use in object and scene observation and will help construct more predictive saliency models in the future.

 \noindent  \\

\noindent \textbf{Acknowledgments} 
We would like to thank Prof. Laurent Itti for providing his eye tracking equipment. We would also like to thank reviewers for their helpful comments on an earlier version of this manuscript.
Please refer to first author's homepage for data and code:  http://pantherfile.uwm.edu/borji/www/.

 \noindent  \\

\bibliographystyle{IEEEtran}
\bibliography{paper2}

\begin{thebibliography}{10}
\providecommand{\url}[1]{#1}
\csname url@rmstyle\endcsname
\providecommand{\newblock}{\relax}
\providecommand{\bibinfo}[2]{#2}
\providecommand\BIBentrySTDinterwordspacing{\spaceskip=0pt\relax}
\providecommand\BIBentryALTinterwordstretchfactor{4}
\providecommand\BIBentryALTinterwordspacing{\spaceskip=\fontdimen2\font plus
\BIBentryALTinterwordstretchfactor\fontdimen3\font minus
  \fontdimen4\font\relax}
\providecommand\BIBforeignlanguage[2]{{%
\expandafter\ifx\csname l@#1\endcsname\relax
\typeout{** WARNING: IEEEtran.bst: No hyphenation pattern has been}%
\typeout{** loaded for the language `#1'. Using the pattern for}%
\typeout{** the default language instead.}%
\else
\language=\csname l@#1\endcsname
\fi
#2}}

\bibitem{henderson1993eye}
J.~M. Henderson, ``Eye movement control during visual object processing:
  effects of initial fixation position and semantic constraint.''
  \emph{Canadian Journal of Experimental Psychology}, vol.~47, no.~1, p.~79,
  1993.

\bibitem{nuthmann2010object}
A.~Nuthmann and J.~M. Henderson, ``Object-based attentional selection in scene
  viewing,'' \emph{Journal of vision}, vol.~10, no.~8, 2010.

\bibitem{xu2014predicting}
J.~Xu, M.~Jiang, S.~Wang, M.~S. Kankanhalli, and Q.~Zhao, ``Predicting human
  gaze beyond pixels,'' vol.~14, no.~1, pp. 1--20, 2014.

\bibitem{borji2013state}
A.~Borji and L.~Itti, ``State-of-the-art in visual attention modeling,''
  vol.~35, no.~1, pp. 185--207, 2013.

\bibitem{hayhoe2005eye}
M.~Hayhoe and D.~Ballard, ``Eye movements in natural behavior,'' \emph{Trends
  in cognitive sciences}, vol.~9, no.~4, pp. 188--194, 2005.

\bibitem{itti2001computational}
L.~Itti and C.~Koch, ``Computational modelling of visual attention,''
  \emph{Nature reviews neuroscience}, vol.~2, no.~3, pp. 194--203, 2001.

\bibitem{land2001ways}
M.~F. Land and M.~Hayhoe, ``In what ways do eye movements contribute to
  everyday activities?'' \emph{Vision research}, vol.~41, no.~25, pp.
  3559--3565, 2001.

\bibitem{navalpakkam2005modeling}
V.~Navalpakkam and L.~Itti, ``Modeling the influence of task on attention,''
  \emph{Vision research}, vol.~45, no.~2, pp. 205--231, 2005.

\bibitem{schutz2011eye}
A.~C. Sch{\"u}tz, D.~I. Braun, and K.~R. Gegenfurtner, ``Eye movements and
  perception: A selective review,'' \emph{Journal of vision}, vol.~11, no.~5,
  p.~9, 2011.

\bibitem{tatler2011eye}
B.~W. Tatler, M.~M. Hayhoe, M.~F. Land, and D.~H. Ballard, ``Eye guidance in
  natural vision: Reinterpreting salience,'' \emph{Journal of vision}, vol.~11,
  no.~5, p.~5, 2011.

\bibitem{baluch2011mechanisms}
F.~Baluch and L.~Itti, ``Mechanisms of top-down attention,'' \emph{Trends in
  neurosciences}, vol.~34, no.~4, pp. 210--224, 2011.

\bibitem{borji2012boosting}
A.~Borji, ``Boosting bottom-up and top-down visual features for saliency
  estimation,'' in \emph{Computer Vision and Pattern Recognition (CVPR), 2012
  IEEE Conference on}.\hskip 1em plus 0.5em minus 0.4em\relax IEEE, 2012, pp.
  438--445.

\bibitem{borji2014salient}
A.~Borji, M.-M. Cheng, H.~Jiang, and J.~Li, ``Salient object detection: A
  survey,'' \emph{arXiv preprint arXiv:1411.5878}, 2014.

\bibitem{BorjiTIP}
A.~Borji, D.~N. Sihite, and L.~Itti, ``Quantitative analysis of human-model
  agreement in visual saliency modeling: A comparative study.'' \emph{IEEE
  Trans. Image Processing.}, vol.~22, no.~1, pp. 55--69, 2012.

\bibitem{2009Cerf}
M.~Cerf, E.~P. Frady, and C.~Koch, ``Faces and text attract gaze independent of
  the task: Experimental data and computer model,'' \emph{Journal of Vision},
  vol.~9, no.~12, November 18 2009.

\bibitem{judd2009learning}
T.~Judd, K.~Ehinger, F.~Durand, and A.~Torralba, ``Learning to predict where
  humans look,'' 2009, pp. 2106--2113.

\bibitem{humphrey2010potency}
K.~Humphrey and G.~Underwood, ``The potency of people in pictures: Evidence
  from sequences of eye fixations,'' \emph{Journal of Vision}, vol.~10, no.~10,
  2010.

\bibitem{wang2012attraction}
H.-C. Wang and M.~Pomplun, ``The attraction of visual attention to texts in
  real-world scenes,'' \emph{Journal of vision}, vol.~12, no.~6, p.~26, 2012.

\bibitem{tatler2007central}
B.~W. Tatler, ``The central fixation bias in scene viewing: Selecting an
  optimal viewing position independently of motor biases and image feature
  distributions,'' \emph{Journal of Vision}, vol.~7, no.~14, p.~4, 2007.

\bibitem{tseng2009quantifying}
P.-H. Tseng, R.~Carmi, I.~G. Cameron, D.~P. Munoz, and L.~Itti, ``Quantifying
  center bias of observers in free viewing of dynamic natural scenes,''
  \emph{Journal of vision}, vol.~9, no.~7, p.~4, 2009.

\bibitem{ossandon2014spatial}
J.~P. Ossand{\'o}n, S.~Onat, and P.~K{\"o}nig, ``Spatial biases in viewing
  behavior,'' \emph{Journal of Vision}, vol.~14, no.~2, p.~20, 2014.

\bibitem{hwang2011semantic}
A.~D. Hwang, H.-C. Wang, and M.~Pomplun, ``Semantic guidance of eye movements
  in real-world scenes,'' \emph{Vision research}, vol.~51, no.~10, pp.
  1192--1205, 2011.

\bibitem{2006Torralba}
A.~Torralba, A.~Oliva, M.~S. Castelhano, and J.~M. Henderson, ``Contextual
  guidance of eye movements and attention in real-world scenes: the role of
  global features in object search,'' \emph{Psychological review}, vol. 113,
  no.~4, pp. 766--786, Oct 2006.

\bibitem{subramanianemotion}
R.~Subramanian, D.~Shankar, N.~Sebe, and D.~Melcher, ``Emotion modulates eye
  movement patterns and subse-quent memory for the gist and details of movie
  scenes,'' 2014.

\bibitem{2005Droll}
J.~A. Droll, M.~M. Hayhoe, J.~Triesch, and B.~T. Sullivan, ``{Task Demands
  Control Acquisition and Storage of Visual Information.}'' \emph{Journal of
  Experimental Psychology Human Perception and Performance}, vol.~31, no.~6,
  pp. 1416--1438, 2005.

\bibitem{carmi2006role}
R.~Carmi and L.~Itti, ``The role of memory in guiding attention during natural
  vision,'' \emph{Journal of Vision}, vol.~6, no.~9, p.~4, 2006.

\bibitem{castelhano2007see}
M.~S. Castelhano, M.~Wieth, and J.~M. Henderson, ``I see what you see: Eye
  movements in real-world scenes are affected by perceived direction of gaze,''
  in \emph{Attention in cognitive systems. Theories and systems from an
  interdisciplinary viewpoint}, 2007, pp. 251--262.

\bibitem{borji2014gaze}
A.~Borji, D.~Parks, and L.~Itti, ``Complementary effects of gaze direction and
  early saliency in guiding fixations during free-viewing,'' \emph{Journal of
  Vision}, vol.~14, no.~13, 2014.

\bibitem{chua2005cultural}
H.~F. Chua, J.~E. Boland, and R.~E. Nisbett, ``Cultural variation in eye
  movements during scene perception,'' \emph{Proceedings of the National
  Academy of Sciences of the United States of America}, vol. 102, no.~35, pp.
  12\,629--12\,633, 2005.

\bibitem{friston1994value}
K.~Friston, G.~Tononi, G.~Reeke~Jr, O.~Sporns, and G.~M. Edelman,
  ``Value-dependent selection in the brain: simulation in a synthetic neural
  model,'' \emph{Neuroscience}, vol.~59, no.~2, pp. 229--243, 1994.

\bibitem{shen2012top}
J.~Shen and L.~Itti, ``Top-down influences on visual attention during listening
  are modulated by observer sex,'' \emph{Vision research}, vol.~65, pp. 62--76,
  2012.

\bibitem{Yarbus1967}
A.~Yarbus, \emph{Eye movements and vision.}\hskip 1em plus 0.5em minus
  0.4em\relax New York: Plenum., 1967.

\bibitem{Land1994}
M.~F. Land and D.~N. Lee, ``Where we look when we steer.'' \emph{Nature}, vol.
  369, pp. 742--744, 1994.

\bibitem{Ballard1995}
D.~Ballard, M.~Hayhoe, and J.~Pelz, ``Memory representations in natural
  tasks.'' \emph{Journal of Cognitive Neuroscience.}, vol.~7, no.~1, pp.
  66--80, 1995.

\bibitem{2001Land}
M.~F. Land and M.~Hayhoe, ``In what ways do eye movements contribute to
  everyday activities?'' \emph{Vision research}, vol.~41, no. 25-26, pp.
  3559--3565, 12 2001.

\bibitem{Borji_etal13smc}
A.~Borji, D.~Sihite, and L.~Itti, ``What/where to look next? modeling top-down
  visual attention in complex interactive environments,'' \emph{IEEE
  Transactions on Systems, Man, and Cybernetics, PART A-SYSTEMS AND HUMANS},
  2014.

\bibitem{borji2014yarbus}
A.~Borji and L.~Itti, ``Defending yarbus: Eye movements predict observers'
  task,'' \emph{Journal of vision}, 2014.

\bibitem{borji2015eyes}
A.~Borji, A.~Lennartz, and M.~Pomplun, ``What do eyes reveal about the mind?:
  Algorithmic inference of search targets from fixations,''
  \emph{Neurocomputing}, vol. 149, pp. 788--799, 2015.

\bibitem{Hajimirza2012}
S.~Hajimirza, M.~Proulx, and E.~Izquierdo, ``Reading users' minds from their
  eyes: A method for implicit image annotation,'' \emph{IEEE Transactions on
  Multimedia}, vol.~14, no.~3, p. 805—815, 2012.

\bibitem{TriesmanGelade}
A.~Treisman and G.~Gelade, ``A feature integration theory of attention.''
  \emph{Cognitive Psychology.}, vol.~12, pp. 97--136, 1980.

\bibitem{Koch_Ullman85}
C.~Koch and S.~Ullman, ``Shifts in selective visual attention: Towards the
  underlying neural circuitry,'' \emph{Human Neurobiology}, vol.~4, no.~4, pp.
  219--227, 1985.

\bibitem{1998Itti}
L.~Itti, C.~Koch, and E.~Niebur, ``A model of saliency-based visual attention
  for rapid scene analysis,'' \emph{IEEE Transactions on Pattern Analysis and
  Machine Intelligence}, vol.~20, no.~11, pp. 1254--1259, Nov 1998.

\bibitem{ReinagelZador99}
P.~Reinagel and A.~Zador, ``Natural scenes at the center of gaze.''
  \emph{Network.}, vol.~10, pp. 341--50, 1999.

\bibitem{ParkhustEtal2002}
D.~Parkhurst, K.~Law, and E.~Niebur, ``Modeling the role of salience in the
  allocation of overt visual attention.'' \emph{Vision Research.}, vol.~42,
  no.~1, pp. 107--123, 2002.

\bibitem{borji2012exploiting}
A.~Borji and L.~Itti, ``Exploiting local and global patch rarities for saliency
  detection,'' in \emph{Computer Vision and Pattern Recognition (CVPR), 2012
  IEEE Conference on}.\hskip 1em plus 0.5em minus 0.4em\relax IEEE, 2012, pp.
  478--485.

\bibitem{hayhoe2003eye}
M.~Hayhoe, A.~Shrivastava, R.~Mruczek, and J.~Pelz, ``Visual memory and motor
  planning in a natural task,'' \emph{Journal of Vision}, vol.~3, p. 49—63,
  2003.

\bibitem{Einhauser2008objects}
W.~Einh{\"a}user, M.~Spain, and P.~Perona, ``Objects predict fixations better
  than early saliency,'' 2008.

\bibitem{borji2013objects}
A.~Borji, D.~N. Sihite, and L.~Itti, ``Objects do not predict fixations better
  than early saliency: A re-analysis of einh{\"a}user et al.'s data,''
  \emph{Journal of vision}, vol.~13, no.~10, p.~18, 2013.

\bibitem{trukenbrod2007oculomotor}
H.~A. Trukenbrod and R.~Engbert, ``Oculomotor control in a sequential search
  task,'' \emph{Vision research}, vol.~47, no.~18, pp. 2426--2443, 2007.

\bibitem{pajak2013object}
M.~Pajak and A.~Nuthmann, ``Object-based saccadic selection during scene
  perception: evidence from viewing position effects,'' \emph{Journal of
  vision}, vol.~13, no.~5, p.~2, 2013.

\bibitem{rensink2000dynamic}
R.~A. Rensink, ``The dynamic representation of scenes,'' \emph{Visual
  cognition}, vol.~7, no. 1-3, pp. 17--42, 2000.

\bibitem{rayner2009rayner}
K.~Rayner, S.~P. Liversedge, A.~Nuthmann, R.~Kliegl, and G.~Underwood,
  ``Rayner's 1979 paper,'' \emph{Perception}, vol.~38, no.~6, p. 895, 2009.

\bibitem{Elazary_Itti08jov}
L.~Elazary and L.~Itti, ``Interesting objects are visually salient,''
  \emph{Journal of Vision}, vol.~8, no. 3:3, pp. 1--15, Mar 2008.

\bibitem{belardinelligaze}
A.~Belardinelli and M.~V. Butz, ``Gaze strategies in object identification and
  manipulation,'' 2013.

\bibitem{liu2011learning}
T.~Liu, Z.~Yuan, J.~Sun, J.~Wang, N.~Zheng, X.~Tang, and H.-Y. Shum, ``Learning
  to detect a salient object,'' \emph{Pattern Analysis and Machine
  Intelligence, IEEE Transactions on}, vol.~33, no.~2, pp. 353--367, 2011.

\bibitem{dziemiankoobject}
M.~Dziemianko, A.~Clarke, and F.~Keller, ``Object-based saliency as a predictor
  of attention in visual tasks.''

\bibitem{borji2013stands}
A.~Borji, D.~N. Sihite, and L.~Itti, ``What stands out in a scene? a study of
  human explicit saliency judgment,'' \emph{Vision research}, vol.~91, pp.
  62--77, 2013.

\bibitem{yun2013exploring}
K.~Yun, Y.~Peng, D.~Samaras, G.~J. Zelinsky, and T.~L. Berg, ``Exploring the
  role of gaze behavior and object detection in scene understanding,''
  \emph{Frontiers in psychology}, vol.~4, 2013.

\bibitem{sun2008computer}
Y.~Sun, R.~Fisher, F.~Wang, and H.~M. Gomes, ``A computer vision model for
  visual-object-based attention and eye movements,'' \emph{Computer Vision and
  Image Understanding}, vol. 112, no.~2, pp. 126--142, 2008.

\bibitem{chang2011fusing}
K.-Y. Chang, T.-L. Liu, H.-T. Chen, and S.-H. Lai, ``Fusing generic objectness
  and visual saliency for salient object detection,'' in \emph{Computer Vision
  (ICCV), 2011 IEEE International Conference on}.\hskip 1em plus 0.5em minus
  0.4em\relax IEEE, 2011, pp. 914--921.

\bibitem{m2013fixations}
B.~M't~Hart, H.~C. Schmidt, C.~Roth, and W.~Einh{\"a}user, ``Fixations on
  objects in natural scenes: dissociating importance from salience,''
  \emph{Frontiers in psychology}, vol.~4, 2013.

\bibitem{kavak2013visual}
Y.~Kavak, E.~Erdem, and A.~Erdem, ``Visual saliency estimation by integrating
  features using multiple kernel learning,'' \emph{arXiv preprint
  arXiv:1307.5693}, 2013.

\bibitem{yanulevskaya2013proto}
V.~Yanulevskaya, J.~Uijlings, J.-M. Geusebroek, N.~Sebe, and A.~Smeulders, ``A
  proto-object-based computational model for visual saliency,'' \emph{Journal
  of vision}, vol.~13, no.~13, p.~27, 2013.

\bibitem{stoll2015overt}
J.~Stoll, M.~Thrun, A.~Nuthmann, and W.~Einh{\"a}user, ``Overt attention in
  natural scenes: Objects dominate features,'' \emph{Vision research}, vol.
  107, pp. 36--48, 2015.

\bibitem{GarciaDiazJOV}
A.~Garcia-Diaz, V.~Leboran, X.~R. Fdez-Vidal, and X.~M. Pardo, ``On the
  relationship between optical variability, visual saliency, and eye fixations:
  A computational approach.'' \emph{Journal of Vision.}, vol.~12, no.~6, 2012.

\bibitem{Peters_etal05vr}
R.~J. Peters, A.~Iyer, L.~Itti, and C.~Koch, ``Components of bottom-up gaze
  allocation in natural images,'' \emph{Vision Research}, vol.~45, no.~8, pp.
  2397--2416, Aug 2005.

\bibitem{hartendorp2013relation}
M.~O. Hartendorp, S.~Van~der Stigchel, I.~Hooge, J.~Mostert, T.~de~Boer, and
  A.~Postma, ``The relation between gaze behavior and categorization: Does
  where we look determine what we see?'' \emph{Journal of vision}, vol.~13,
  no.~6, p.~6, 2013.

\bibitem{kootstra2011predicting}
G.~Kootstra, B.~de~Boer, and L.~R. Schomaker, ``Predicting eye fixations on
  complex visual stimuli using local symmetry,'' \emph{Cognitive computation},
  vol.~3, no.~1, pp. 223--240, 2011.

\bibitem{greene2013statistics}
M.~R. Greene, ``Statistics of high-level scene context,'' \emph{Frontiers in
  psychology}, vol.~4, 2013.

\bibitem{cheng2014salientshape}
M.-M. Cheng, N.~J. Mitra, X.~Huang, and S.-M. Hu, ``Salientshape: Group
  saliency in image collections,'' \emph{The Visual Computer}, vol.~30, no.~4,
  pp. 443--453, 2014.

\bibitem{li2014secrets}
Y.~Li, X.~Hou, C.~Koch, J.~Rehg, and A.~Yuille, ``The secrets of salient object
  segmentation.''\hskip 1em plus 0.5em minus 0.4em\relax CVPR, 2014.

\bibitem{duncan1984selective}
J.~Duncan, ``Selective attention and the organization of visual information.''
  \emph{Journal of Experimental Psychology: General}, vol. 113, no.~4, p. 501,
  1984.

\bibitem{egly1994shifting}
R.~Egly, J.~Driver, and R.~D. Rafal, ``Shifting visual attention between
  objects and locations: evidence from normal and parietal lesion subjects.''
  \emph{Journal of Experimental Psychology: General}, vol. 123, no.~2, p. 161,
  1994.

\bibitem{vecera1994does}
S.~P. Vecera and M.~J. Farah, ``Does visual attention select objects or
  locations?'' \emph{Journal of Experimental Psychology: General}, vol. 123,
  no.~2, p. 146, 1994.

\bibitem{drummond2010object}
L.~Drummond and S.~Shomstein, ``Object-based attention: Shifting or
  uncertainty?'' \emph{Attention, Perception, \& Psychophysics}, vol.~72,
  no.~7, pp. 1743--1755, 2010.

\bibitem{gottlieb2010attention}
J.~Gottlieb and P.~Balan, ``Attention as a decision in information space,''
  \emph{Trends in cognitive sciences}, vol.~14, no.~6, pp. 240--248, 2010.

\bibitem{harel2007graph}
J.~Harel, C.~Koch, P.~Perona, \emph{et~al.}, ``Graph-based visual saliency,''
  \emph{Advances in neural information processing systems}, vol.~19, p. 545,
  2007.

\bibitem{koehler2014saliency}
K.~Koehler, F.~Guo, S.~Zhang, and M.~P. Eckstein, ``What do saliency models
  predict?'' \emph{Journal of vision}, vol.~14, no.~3, p.~14, 2014.

\bibitem{YiHou2014}
Y.~Li, X.~Hou, C.~Koch, J.~Rehg, and A.~Yuille, ``The secrets of salient object
  segmentation.''\hskip 1em plus 0.5em minus 0.4em\relax CVPR, 2014.

\bibitem{everingham2010pascal}
M.~Everingham, L.~Van~Gool, C.~K. Williams, J.~Winn, and A.~Zisserman, ``The
  pascal visual object classes (voc) challenge,'' \emph{International journal
  of computer vision}, vol.~88, no.~2, pp. 303--338, 2010.

\end{thebibliography}

%

%
%

\vspace{-50pt}

\begin{biography}[{\includegraphics[width=1in,height=1.25in,clip,keepaspectratio]{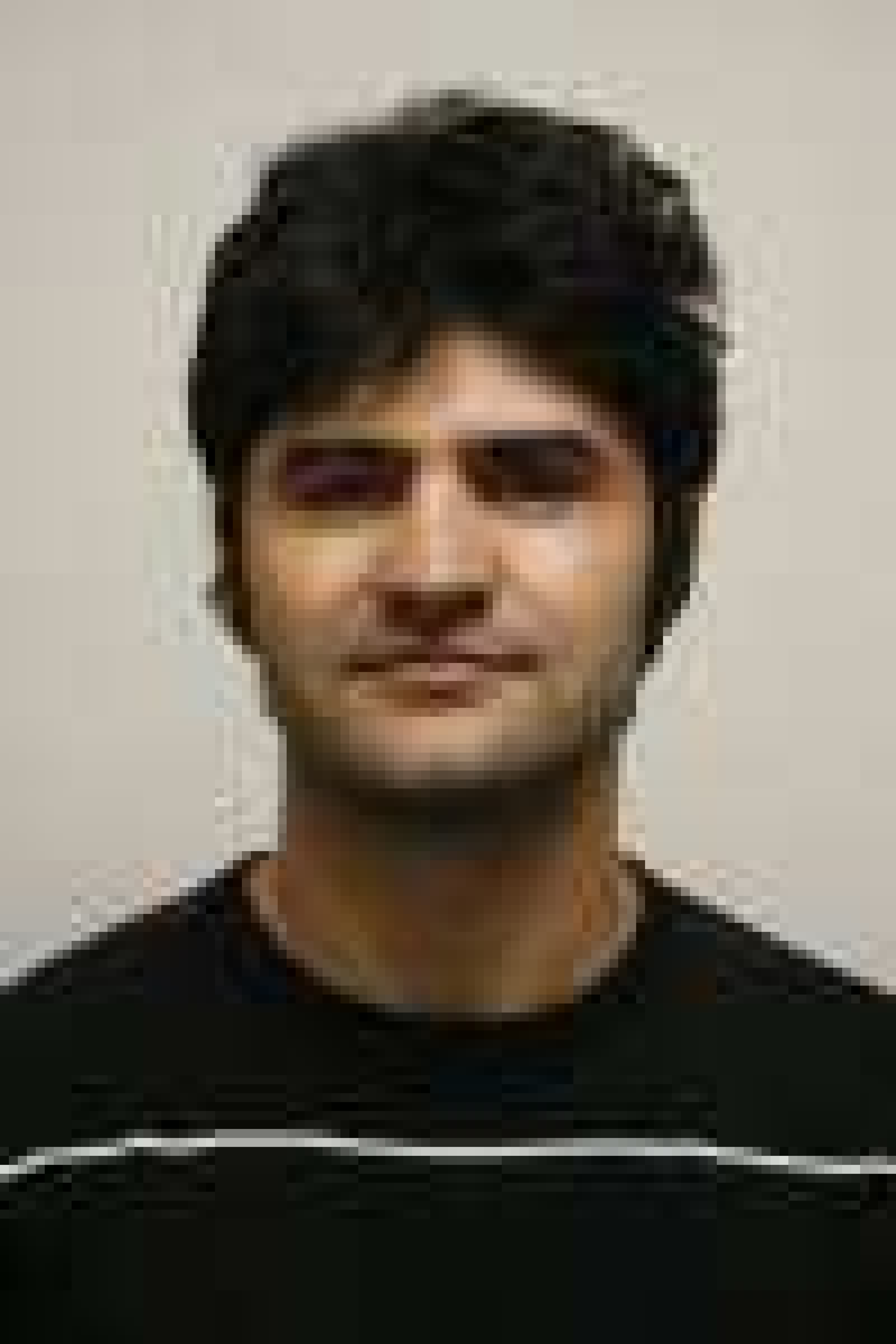}}]{Ali Borji}
received his BS and MS degrees in computer engineering from Petroleum University of Technology, Tehran, Iran, 2001 and Shiraz University, Shiraz, Iran, 2004, respectively. He did his Ph.D. in cognitive neurosciences at Institute for Studies in Fundamental Sciences (IPM) in Tehran, Iran, 2009 and spent four years as a postdoctoral scholar at iLab, University of Southern California from 2010 to 2014. He is currently an assistant professor at University of Wisconsin, Milwaukee. His research interests include visual attention, active learning, object and scene recognition, and cognitive and computational neurosciences.
\end{biography}

\vspace{-50pt}

\begin{IEEEbiography}[{\includegraphics[width=1in,height=1.25in, clip,keepaspectratio]{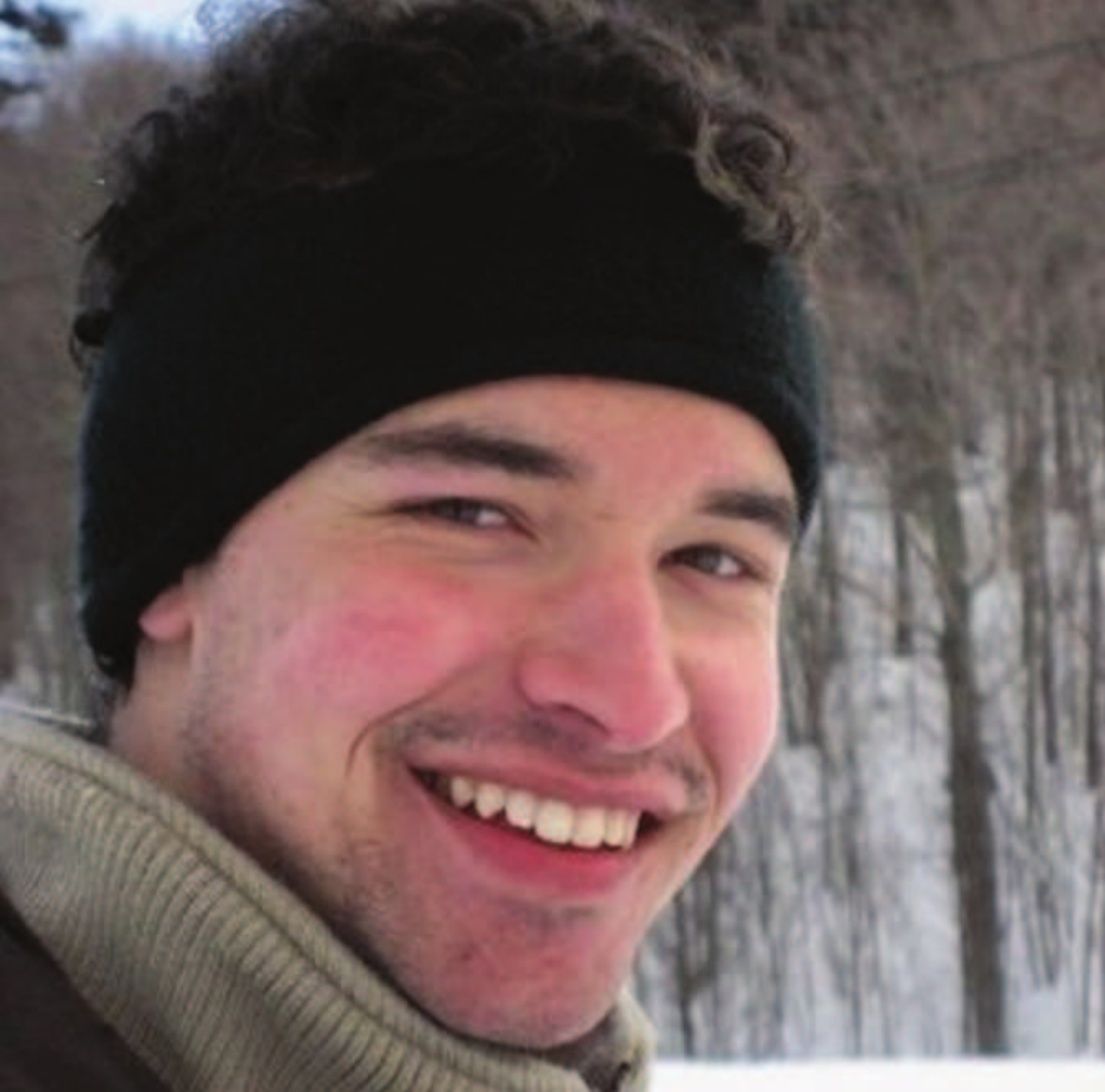}}]{James Tanner} earned a B.S. in Computer Science and Mathematics from the University of Maryland, and currently is a doctoral student in Department of Computer Science at the University of Southern California. His interests include computer vision, machine learning, and neural networks.
\end{IEEEbiography}

\end{document}